\documentclass{article}

\usepackage[usenames,dvipsnames]{xcolor}
\usepackage{url}            %
\usepackage{booktabs}       %
\usepackage{amsfonts}       %
\usepackage{nicefrac}       %
\usepackage{microtype}      %
\usepackage{amsmath,amssymb,bbm}
\usepackage{textcomp}       %
\usepackage{gensymb}        %
\usepackage{enumitem}       %
\usepackage{multirow}
\usepackage{algorithm}
\usepackage{algorithmic}
\usepackage{subcaption}
\usepackage{hyperref}       %
\usepackage{wrapfig}        %
\usepackage{bbding}         %
\usepackage{mathtools}

\newcommand{\Skip}[1]{}

\renewcommand{\algorithmiccomment}[1]{{\hfill\footnotesize{\color{gray}{$\triangleright$ #1}}}}

\PassOptionsToPackage{numbers, compress}{natbib}

\usepackage[final]{neurips_2023}
\usepackage{booktabs}
\usepackage{multirow}
\usepackage{wrapfig}
\usepackage{lineno}
\usepackage{stmaryrd}

\title{Cross-Episodic Curriculum for Transformer Agents}

\author{
Lucy Xiaoyang Shi$^{1*}$, Yunfan Jiang$^{1*}$, Jake Grigsby$^2$, Linxi Fan$^{3 \dagger}$, Yuke Zhu$^{2\,3 \dagger}$\\
$^1$Stanford University \quad
$^2$The University of Texas at Austin \quad
$^3$NVIDIA Research\\
$^*$Equal contribution \, $^\dagger$Equal advising
}

\usepackage{soul}
\usepackage{xspace}

\definecolor{darkgreen}{rgb}{0.09, 0.45, 0.27}
\newcommand{\para}[1]{\paragraph{#1}\looseness=-1}
\newcommand{\bestscore}[1]{\textcolor{darkgreen}{\mathbf{#1}}}

\newcommand{\cec}[0]{\mbox{CEC}\xspace}

\newcommand{\webpage}[0]{\href{https://cec-agent.github.io/}{\texttt{cec-agent.github.io}}}

\begin{document}
\maketitle

\begin{abstract}
    We present a new algorithm, Cross-Episodic Curriculum (\cec), to boost the learning efficiency and generalization of Transformer agents. 
Central to \cec is the placement of \textit{cross-episodic} experiences into a Transformer’s context, which forms the basis of a curriculum.
By sequentially structuring online learning trials and mixed-quality demonstrations,
\cec constructs curricula that encapsulate learning progression and proficiency increase across episodes.
Such synergy combined with the potent pattern recognition capabilities of Transformer models delivers a powerful \emph{cross-episodic attention} mechanism.
The effectiveness of \cec is demonstrated under two representative scenarios: one involving multi-task reinforcement learning with discrete control, such as in DeepMind Lab, where the curriculum captures the learning progression in both individual and progressively complex settings, and the other involving imitation learning with mixed-quality data for continuous control, as seen in RoboMimic, where the curriculum captures the improvement in demonstrators' expertise.
In all instances, policies resulting from \cec exhibit superior performance and strong generalization.
Code is open-sourced on the project website \webpage{} to facilitate research on Transformer agent learning.

\end{abstract}

\section{Introduction}
\label{sec:introduction}

The paradigm shift driven by foundation models~\citep{bommasani2021opportunities} is revolutionizing the communities who study sequential decision-making problems~\citep{yang2023foundation}, with innovations focusing on control~\citep{ahn2022saycan,liang2022code,jiang2022vima,brohan2022rt1}, planning~\citep{wang2023voyager,huang2022language,huang2022inner,wang2023describe,driess2023palme}, pre-trained visual representation~\citep{nair2022r3m,ma2022vip,radosavovic2022realworld,majumdar2023search}, among others.
Despite the progress, the data-hungry nature makes the application of Transformer~\citep{vaswani2017attention} agents extremely challenging in data-scarce domains like robotics~\citep{mandlekar2018roboturk,mandlekar2021matters,ebert2021bridge,jiang2022vima,brohan2022rt1}.
This leads us to the question: Can we maximize the utilization of limited data, regardless of their optimality and construction, to foster more efficient learning?

\begin{figure}[!t]
    \centering
    \vspace{-0.1in}
    \makebox[\textwidth][c]{\includegraphics[width=0.9\textwidth]{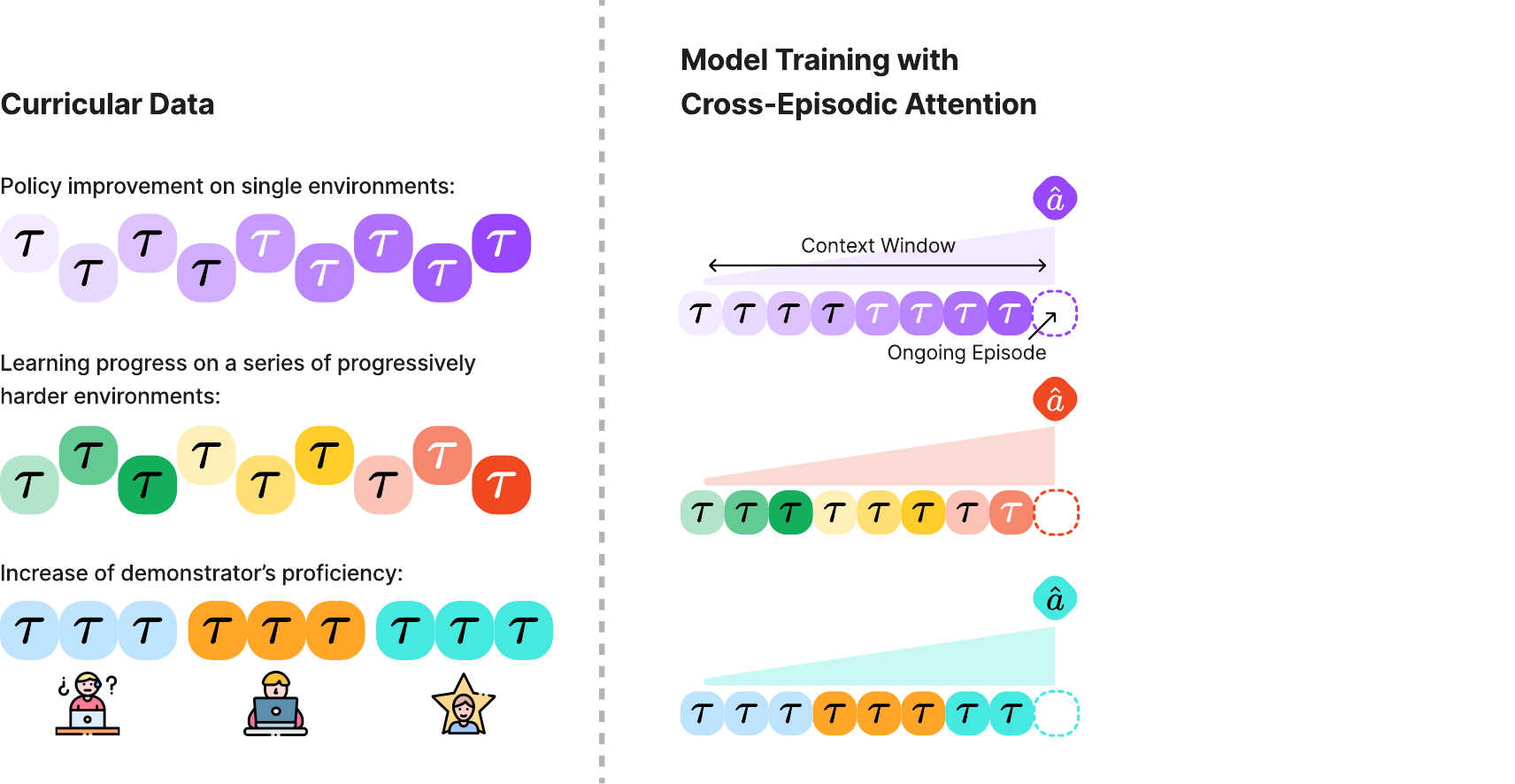}}
    \caption{\textbf{Cross-episodic curriculum for Transformer agents.}
    \cec involves two major steps:
    \textit{\mbox{1) Preparation of curricular data.}} We order multiple experiences such that they explicitly capture curricular patterns. For instance, they can be policy improvement in single environments, learning progress in a series of progressively harder environments, or the increase of the demonstrator's expertise.
    \textit{2) Model training with cross-episodic attention.} When training the model to predict actions, it can trace back beyond the current episode and internalize the policy refinement
    for more efficient learning.
    Here each $\tau$ represents an episode (trajectory). $\hat{a}$ refers to actions predicted by the model.
    Colored triangles denote causal Transformer models.
    }
    \label{fig:pull_fig}  
\end{figure}

To this end, this paper introduces a novel algorithm named \textit{Cross-Episodic Curriculum} (\cec), a method that explicitly harnesses the shifting distributions of multiple experiences when organized into a curriculum.
The key insight is that sequential \textit{cross-episodic} data manifest useful learning signals that do not easily appear in any separated training episodes.\footnote{Following the canonical definition in \citet{sutton2018reinforcement}, we refer to the sequences of agent-environment interaction with clearly identified initial and terminal states as ``episodes''. We interchangeably use ``episode'', ``trial'', and ``trajectory'' in this work.}
As illustrated in Figure~\ref{fig:pull_fig}, \cec realizes this through two stages: 1) formulating curricular sequences to capture (a) the policy improvement on single environments, (b) the learning progress on a series of progressively harder environments, or (c) the increase of demonstrators' proficiency; 
and 2) causally distilling policy improvements into the model weights of Transformer agents through \emph{cross-episodic attention}.
When a policy is trained to predict actions at current time steps, it can trace back beyond ongoing trials and internalize improved behaviors encoded in curricular data, thereby achieving efficient learning and robust deployment when probed with visual or dynamics perturbations.
Contrary to prior works like Algorithm Distillation (AD, \citet{laskin2023incontext}) which, at test time, samples and retains a single task configuration across episodes for in-context refinement, our method, \cec, prioritizes zero-shot generalization across a distribution of test configurations. With \cec, agents are evaluated on a new task configuration in each episode, emphasizing adaptability to diverse tasks.

We investigate the effectiveness of \cec in enhancing sample efficiency and generalization with two representative case studies.
They are: 1) Reinforcement Learning (RL) on DeepMind Lab~(DMLab)~\citep{beattie2016deepmind}, a 3D simulation encompassing visually diverse worlds, complicated environment dynamics, ego-centric pixel inputs, and joystick control; and 2) Imitation Learning (IL) from mixed-quality human demonstrations on RoboMimic~\citep{mandlekar2021matters}, a framework designed to study robotic manipulation with proprioceptive and external camera observations and continuous control.
Despite RL episodes being characterized by state-action-reward tuples and IL trajectories by state-action pairs, our method exclusively employs state-action pairs in its approach.

In challenging embodied navigation tasks, despite significant generalization gaps (Table~\ref{table:dmlab_exp_setting}),
our method surpasses concurrent and competitive method Agentic Transformer (AT, \citet{liu2023emergent}).
It also significantly outperforms popular offline RL methods such as Decision Transformer (DT, \citet{chen2021decisiontransformer}) and baselines trained on expert data, with the same amount of parameters, architecture, and data size.
It even exceeds RL oracles directly trained on test task distributions 
by $50\%$ in a \emph{zero-shot} manner. 
\cec also yields robust embodied policies that are up to $1.6\times$ better than RL oracles when zero-shot probed with unseen environment dynamics.
When learning continuous robotic control, \cec successfully solves two simulated manipulation tasks, matching and outperforming previous well-established baselines~\cite {mandlekar2021matters,fujimoto2018offpolicy,kumar2020conservative}.
Further ablation reveals that \cec with cross-episodic attention is a generally effective recipe for learning Transformer agents, especially in applications where sequential data exhibit moderate and smooth progression.

\section{Cross-Episodic Curriculum: Formalism and Implementations}
\label{sec:method}
In this section, we establish the foundation for our cross-episodic curriculum method by first reviewing the preliminaries underlying our case studies, which encompass two representative scenarios in sequential decision-making. Subsequently, we formally introduce the assembly of curricular data and the specifics of model optimization utilizing cross-episodic attention. Lastly, we delve into the practical implementation of \cec in the context of these two scenarios.

\subsection{Preliminaries}
\para{Reinforcement learning.}
We consider the setting where source agents learn through trial and error in partially observable environments.
Denoting states $s \in \mathcal{S}$ and actions $a \in \mathcal{A}$, an agent interacts in a Partially Observable Markov Decision Process (POMDP) with the transition function $p(s_{t + 1} \vert s_t, a_t): \mathcal{S} \times \mathcal{A} \rightarrow \mathcal{S}$. It observes $o \in \mathcal{O}$ emitted from observation function $\Omega(o_t \vert s_t, a_{t - 1}): \mathcal{S} \times \mathcal{A} \rightarrow \mathcal{O}$ and receives scalar reward $r$ from $R (s, a): \mathcal{S} \times \mathcal{A} \rightarrow \mathbb{R}$.
Under the episodic task setting,
RL seeks to learn a parameterized policy $\pi_\theta (\cdot \vert s)$ that maximizes the return over a fixed length $T$ of interaction steps: $\pi_\theta = \arg \max_{\theta \in \Theta} \sum_{t = 0}^{T - 1} \gamma^t r_t$, where $\gamma \in [0, 1)$ is a discount factor.
Here we follow the canonical definition of an episode $\tau$ as a series of environment-agent interactions with length $T$, $\tau \vcentcolon= (s_0, a_0, r_0, \ldots, s_{T - 1}, a_{T - 1}, r_{T - 1}, s_T)$, where initial states $s_0$ are sampled from initial state distribution $s_0 \sim \rho_0(s)$ and terminal states $s_T$ are reached once the elapsed timestep exceeds $T$.
Additionally, we view all RL tasks considered in this work as goal-reaching problems~\citep{Kaelbling1993LearningTA,ghosh2019learning} and constrain all episodes to terminate upon task completion.
It is worth noting that similar to previous work~\citep{laskin2023incontext}, training data are collected by source RL agents during their online learning. Nevertheless, once the dataset is obtained, our method is trained \emph{offline} in a purely supervised manner.

\para{Imitation learning.}
We consider IL settings with existing trajectories composed only of state-action pairs.
Furthermore, we relax the assumption on demonstration optimality and allow them to be crowdsourced~\citep{brown2019extrapolating,chen2020learning,cao2021learning}.
Data collected by operators with varying expertise are therefore unavoidable. 
Formally, we assume the access to a dataset $\mathcal{D}^N \vcentcolon= \{\tau_1, \ldots, \tau_{N}\}$ consisting of $N$ demonstrations, with each demonstrated trajectory $\tau_i \vcentcolon= (s_0, a_0, \ldots, s_{T - 1}, a_{T - 1})$ naturally identified as an episode.
The goal of IL, specifically of behavior cloning (BC), is to learn a policy $\pi_\theta$ that accurately models the distribution of behaviors.
When viewed as goal-reaching problems, BC policies can be evaluated by measuring the success ratio in completing tasks~\citep{ghosh2019learning}.

\subsection{Curricular Data Assembly and Model Optimization}
Meaningful learning signals emerge when multiple trajectories are organized and examined cross-episodically along a curriculum axis.
This valuable information, which is not easily discernible in individual training episodes, may encompass aspects such as the improvement of an RL agent's navigation policy or the generally effective manipulation skills exhibited by operators with diverse proficiency levels.
With a powerful model architecture such as Transformer~\cite {vaswani2017attention,dai2019transformerxl}, such emergent and valuable learning signals can be baked into policy weights, thereby boosting performance in embodied tasks.

For a given embodied task $\mathcal{M}$, we define its curriculum $\mathcal{C}_\mathcal{M}$ as a collection of trajectories $\tau$ consisting of state-action pairs.
A series of ordered levels $[\mathcal{L}_1, \ldots, \mathcal{L}_L]$ partitions this collection such that $\bigcup_{l \in \{1, \ldots, L\}} \mathcal{L}_{l} = \mathcal{C}_\mathcal{M}$ and $\bigcap_{\forall i, j \in \{1, \ldots, L\}, i \neq j} \mathcal{L}_{\{i, j\}} = \emptyset$.
More importantly, these ordered levels characterize a curriculum by encoding, for example, learning progress in single environments, learning progress in a series of progressively harder environments, or the increase of the demonstrator's expertise.

With a curriculum $\mathcal{C}_\mathcal{M} \vcentcolon= \{\tau_i\}_{i = 1}^N$ and its characteristics $[\mathcal{L}_1, \ldots, \mathcal{L}_L]$, we construct a curricular sequence $\mathcal{T}$ that spans multiple episodes and captures the essence of gradual improvement in the following way:
\begin{equation}
    \mathcal{T} \vcentcolon= \bigoplus_{l \in \{1, \ldots, L\}} \left[\tau^{(1)}, \ldots, \tau^{(C)} \right], \quad  \text{where} \quad C \sim  \mathcal{U}\left(\llbracket \vert \mathcal{L}_l \vert \rrbracket\right) \quad \text{and} \quad \tau^{(c)} \sim \mathcal{L}_l.
\end{equation}
The symbol $\oplus$ denotes the concatenation operation. $\mathcal{U}\left( \llbracket K \rrbracket \right)$ denotes a uniform distribution over the discrete set $\{k \in \mathbb{N}, k\leq K \}$. In practice, we use values smaller than $\vert \mathcal{L}_l \vert$ considering the memory consumption.

We subsequently learn a causal policy that only depends on cross-episodic historical observations $\pi_\theta (\cdot \vert o_{\leq t}^{(\leq n)})$.
Note that this modeling strategy differs from previous work that views sequential decision-making as a big sequence-modeling problem~\citep{chen2021decisiontransformer,janner2021onebigsequence,laskin2023incontext,jiang2022vima}. It instead resembles the causal policy in \citet{openai2022vpt}. Nevertheless, we still follow the best practice~\citep{for-the-win,openai2019dota,fan2022minedojo} to provide previous action as an extra modality of observations in POMDP RL tasks.

We leverage the powerful attention mechanism of Transformer~\citep{vaswani2017attention} to enable cross-episodic attention.
Given observation series $O_t^{(n)} \vcentcolon= \{o_{0}^{(1)},\ldots, o_{\leq t}^{(\leq n)} \}$ (shorthanded as $O$ hereafter for brevity), 
Transformer projects it into query $Q = f_Q(O)$, key $K = f_K (O)$, and value $V = f_V(O)$ matrices, with each row being a $D$-dim vector.
Attention operation is performed to aggregate information:
\begin{equation}
    \text{Attention}(Q, K, V) = \text{softmax}(\frac{QK^\intercal}{\sqrt{D}})V.
\end{equation}
Depending on whether the input arguments for $f_Q$ and $f_{\{K, V\}}$ are the same, attention operation can be further divided into self-attention and cross-attention.
Since tasks considered in this work do not require additional conditioning for task specification, 
we follow previous work~\citep{openai2022vpt,zhu2022viola} to utilize self-attention to process observation series.
Nevertheless, ours can be naturally extended to handle, for example, natural language or multi-modal task prompts, following the cross-attention introduced in \citet{jiang2022vima}.

Finally, this Transformer policy is trained by simply minimizing the negative log-likelihood objective $\mathcal{J}_{\text{NLL}}$ of labeled actions, conditioned on cross-episodic context:
\begin{equation}
    \mathcal{J}_{\text{NLL}} =   - \log \pi_\theta (\cdot \vert \mathcal{T}) = \frac{1}{\vert \mathcal{T} \vert \times T} \sum_{n = 1}^{\vert \mathcal{T} \vert} \sum_{t = 1}^T - \log \pi_\theta \left(a_t^{(n)} \vert  o_{\leq t}^{(\leq n)} \right).
\end{equation}
Regarding the specific memory architecture, we follow \citet{openai2022vpt,team2023humantimescale} to use Transformer-XL~\citep{dai2019transformerxl} as our model backbone. Thus, during deployment, we keep its hidden states propagating across test episodes to mimic the training settings.

\subsection{Practical Implementations}
We now discuss concrete instantiations of \cec for 1) RL with DMLab and 2) IL with RoboMimic.
Detailed introductions to the benchmark and task selection are deferred to Sec.~\ref{sec:exp_setup}.
We investigate the following three curricula, where the initial two pertain to RL, while the final one applies to IL:

\para{Learning-progress-based curriculum.}
In the first instantiation, inspired by the literature on learning progress~\citep{matiisen2017teacherstudent,graves2017automated,portelas2019teacher,kanitscheider2021multitask}, we view the progression of learning agents as a curriculum.
Concretely, we train multi-task PPO agents~\citep{schulman2017proximal,petrenko2020sample} on tasks drawn from test distributions.
We record their online interactions during training, which faithfully reflect the learning progress.
Finally, we form the \textit{learning-progress-based curriculum} by sequentially concatenating episodes collected at different learning stages.
Note that this procedure is different from \citet{laskin2023incontext}, where for each environment, the learning dynamics of \emph{multiple} single-task RL agents has to be logged. In contrast, we only track a \emph{single} multi-task agent per environment.
\begin{figure}[t]
    \centering
    \begin{subfigure}[t]{0.19\textwidth}
        \centering
        \includegraphics[width=\textwidth]{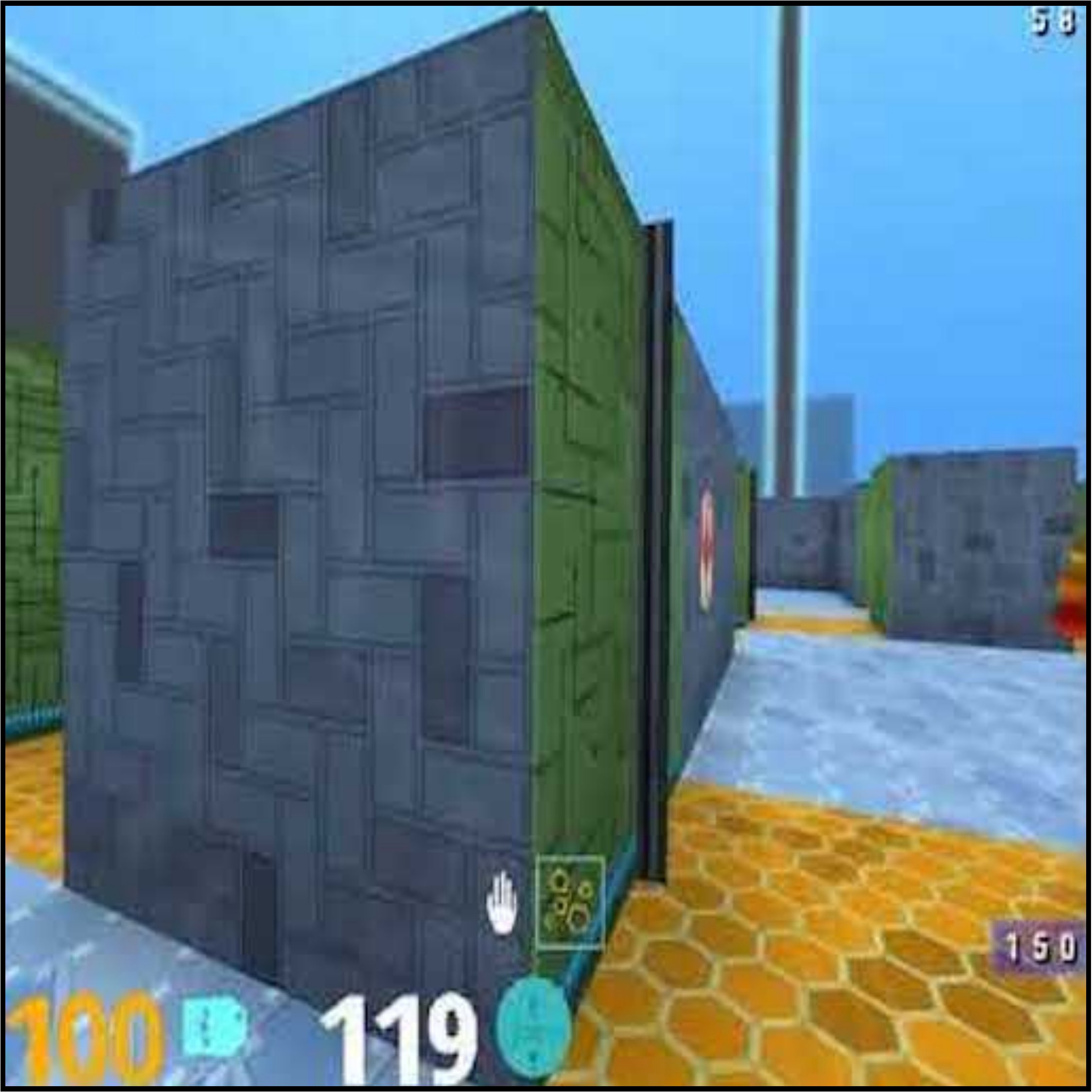}
        \caption{Goal Maze}
        \label{fig:tasks:goal_maze}
    \end{subfigure} 
    \hfill
    \begin{subfigure}[t]{0.19\textwidth}
        \centering
        \includegraphics[width=\textwidth]{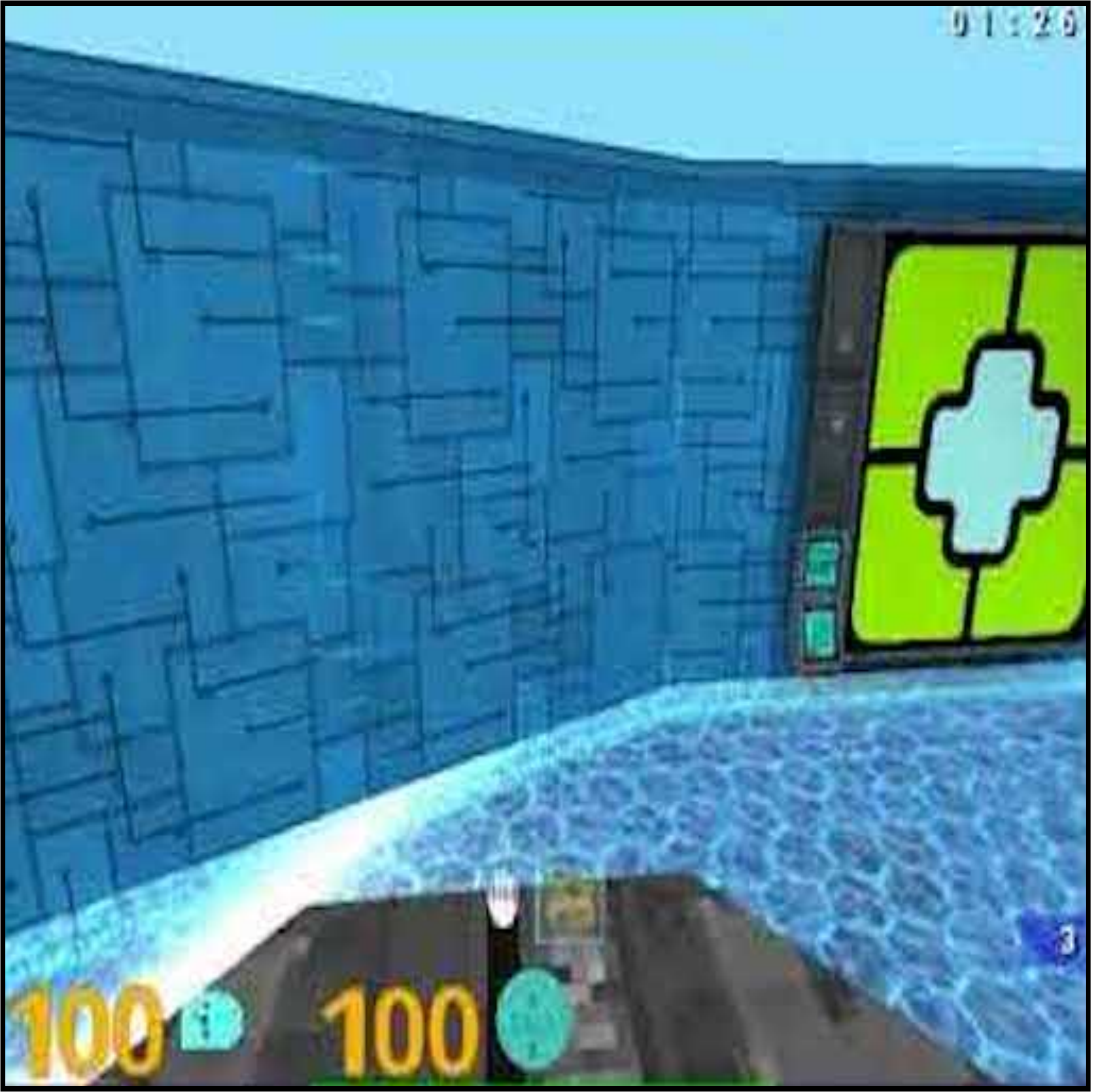}
        \caption{Watermaze}
        \label{fig:tasks:watermaze}
    \end{subfigure}
    \hfill
    \begin{subfigure}[t]{0.19\linewidth}
        \centering
        \includegraphics[width=\linewidth]{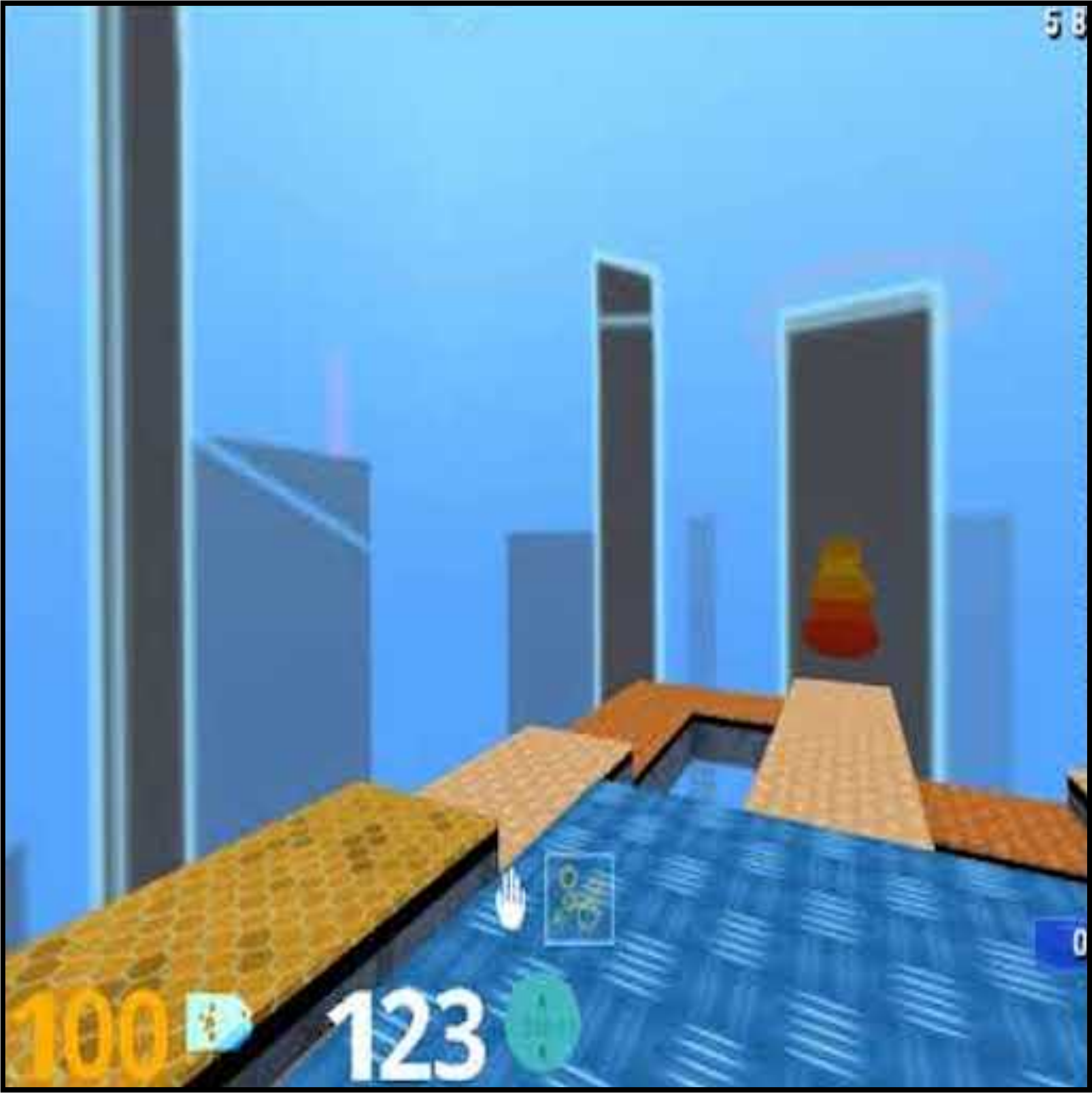}
        \caption{Irreversible Path}
        \label{fig:tasks:irreversible_path}
    \end{subfigure}
    \hfill
    \begin{subfigure}[t]{0.19\linewidth}
        \centering
        \includegraphics[width=\linewidth]{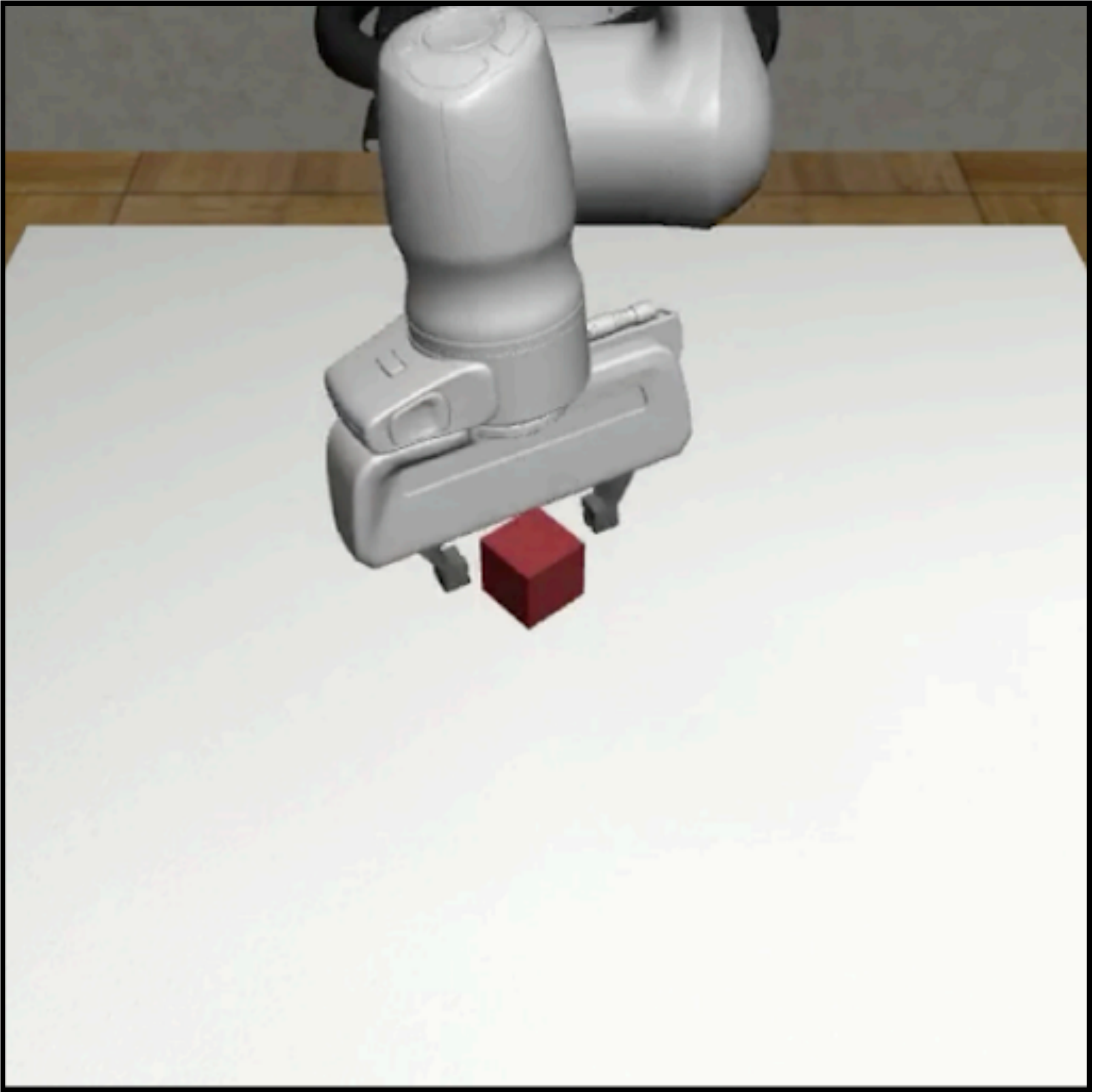}
        \caption{Lift}
        \label{fig:tasks:lift}
    \end{subfigure} 
    \hfill
    \begin{subfigure}[t]{0.19\linewidth}
        \centering
        \includegraphics[width=\linewidth]{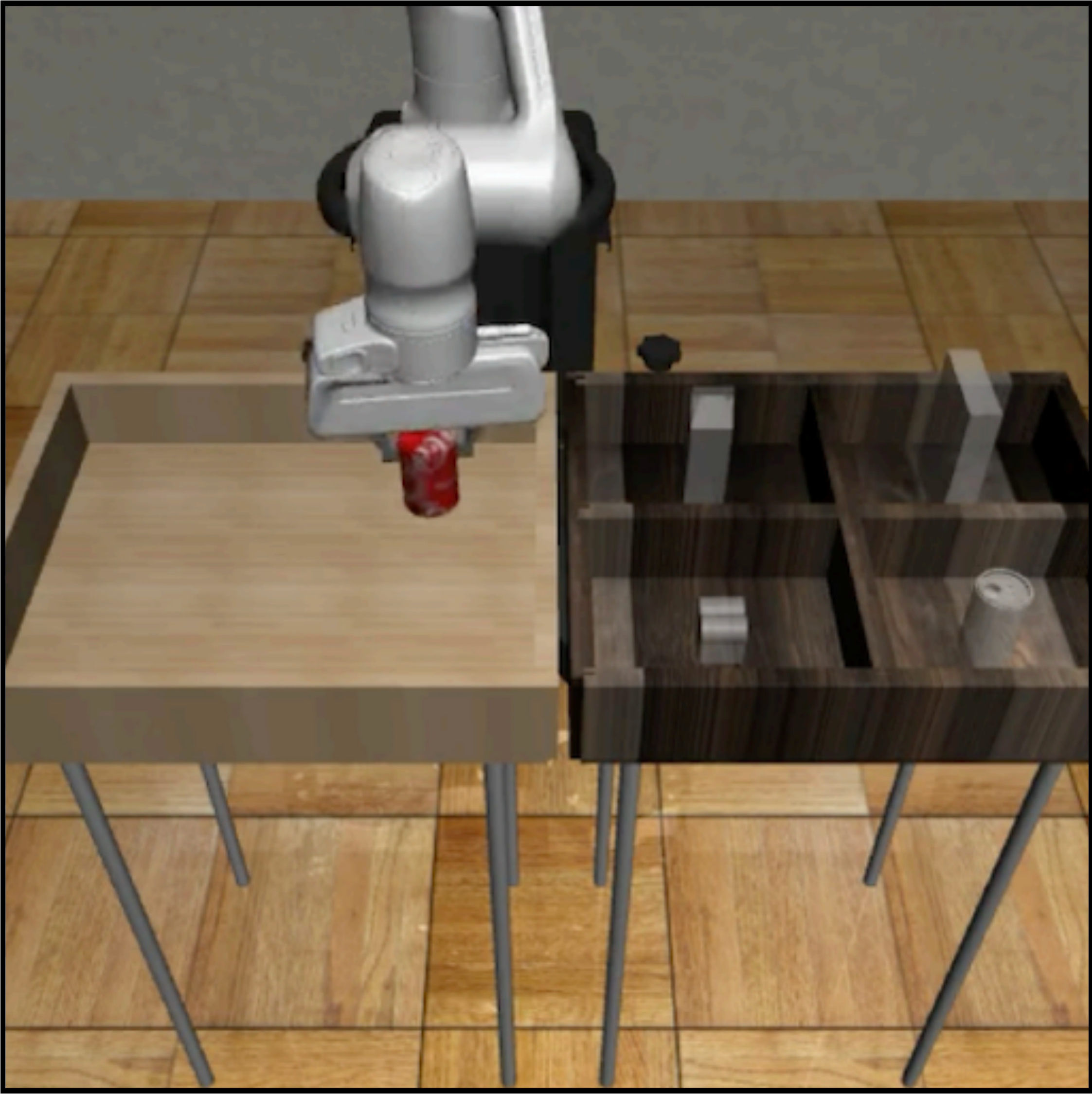}
        \caption{Can}
        \label{fig:tasks:can}
    \end{subfigure} 
    \caption{
        We evaluate our method on five tasks that cover challenges such as exploration and planning over long horizons in RL settings, as well as object manipulation and continuous control in IL settings.
        Figures are from \citet{beattie2016deepmind} and \citet{mandlekar2021matters}.
    }
    \label{fig:tasks}
  \vspace{-1em}
\end{figure}

\para{Task-difficulty-based curriculum.} In the second instantiation, instead of taking snapshots of RL agents directly trained on test configurations, we collect learning progress on a series of easier but progressively harder tasks. For instance, in an embodied navigation task, the test configuration includes 20 rooms. Rather than logging source agents' learning progression in the 20-room maze, we record in a series of mazes with 5, 10, and 15 rooms. We then structure stored episodes first following learning progress and then the increase of layout complexity. This practice naturally creates a \textit{task-difficulty-based curriculum}, which resembles curriculum RL that is based on task difficulty ~\citep{matiisen2017teacherstudent,Narvekar2017AutonomousTS}.
We find it especially helpful for hard-exploration problems where the source RL agent does not make meaningful progress.

\para{Expertise-based curriculum.} For the setting of IL from mixed-quality demonstrations, we instantiate a curriculum based on demonstrators' expertise. This design choice is motivated by literature on learning from heterogeneous demonstrators~\citep{beliaev2022imitation, NEURIPS2021_670e8a43}, with the intuition that there is little to learn from novices but a lot from experts. To realize this idea, we leverage the Multi-Human dataset from RoboMimic~\citep{mandlekar2021matters}. Since it contains demonstrations collected by human demonstrators with varying proficiency, we organize offline demonstration trajectories following the increase of expertise to construct the \textit{expertise-based curriculum}.

\section{Experimental Setup}
\label{sec:exp_setup}

In this section, we elaborate on the experimental setup of our case studies. Our investigation spans two representative and distinct settings:
1) online reinforcement learning with 3D maze environments of DMLab~\citep{beattie2016deepmind}, and 2) imitation learning from mixed-quality human demonstrations of RoboMimic~\citep{mandlekar2021matters}. For each of them, we discuss task selection, baselines, and training and evaluation protocols.
Teasers of these tasks are shown in Figure~\ref{fig:tasks}.

\subsection{Task Settings and Environments}

\textbf{DeepMind Lab}~\citep{beattie2016deepmind} is a 3D learning environment with diverse tasks.
Agents spawn in visually complex worlds, receive ego-centric (thus partially observable) RGB pixel inputs, and execute joystick actions. We consider three levels from this benchmark: Goal Maze, Watermaze~\citep{watermaze}, and Sky Maze with Irreversible Path. They challenge agents to explore, memorize, and plan over a long horizon. Their goals are similar --- to navigate in complicated mazes and find a randomly spawned goal, upon which sparse rewards will be released.
Episodes start with randomly spawned agents and goals and terminate once goals are reached or elapsed steps have exceeded pre-defined horizons.

\textbf{RoboMimic}~\citep{mandlekar2021matters} is a framework designed for studying robot manipulation and learning from demonstrations. Agents control robot arms with fixed bases, receive proprioceptive measurements and image observations from mounted cameras, and operate with continuous control.
We evaluate two simulated tasks: ``Lift'' and ``Can''. In the ``Lift'' task, robots are tasked with picking up a small cube. In the ``Can'' task, robots are required to pick up a soda can from a large bin and place it into a smaller target bin.
Episodes start with randomly initialized object configuration and terminate upon successfully completing the task or exceeding pre-defined horizons.

\begin{table}[t!]
\caption{\textbf{Generalization gaps between training and testing for DMLab levels.} Note that agents resulting from task-difficulty-based curricula are not trained on test configurations. Therefore, their performance should be considered as \emph{zero-shot}.}
\label{table:dmlab_exp_setting}
\vspace{0.1in}
\centering
\resizebox{1\textwidth}{!}{
\begin{tabular}{@{}c|c|c|ccccc@{}}
\toprule
\multirow{2}{*}{\textbf{\begin{tabular}[c]{@{}c@{}}Level\\ Name\end{tabular}}} & \multirow{2}{*}{\textbf{\begin{tabular}[c]{@{}c@{}}Difficulty\\ Parameter\end{tabular}}} & \multirow{2}{*}{\textbf{\begin{tabular}[c]{@{}c@{}}Test\\ Difficulty\end{tabular}}} & \multicolumn{5}{c}{\textbf{Training Difficulty}}                                                                                                                                                                                                                                                                               \\ \cmidrule(l){4-8} 
                                                                               &                                                                                          &                                                                                     & \begin{tabular}[c]{@{}c@{}}Ours\\ (Learning Progress)\end{tabular} & \begin{tabular}[c]{@{}c@{}}Ours\\ (Task Difficulty)\end{tabular} & \begin{tabular}[c]{@{}c@{}}BC\\ w/ Expert Data\end{tabular} & \begin{tabular}[c]{@{}c@{}}RL\\ (Oracle)\end{tabular} & \begin{tabular}[c]{@{}c@{}}Curriculum RL\\ (Oracle)\end{tabular} \\ \midrule
Goal Maze                                                                      & Room Numbers                                                                             & 20                                                                                  & 20                                                                 & 5→10→15                                                          & 20                                                          & 20                                                    & 5→10→15→20                                                       \\
Watermaze                                                                      & Spawn Radius                                                                             & 580                                                                                 & 580                                                                & 150→300→450                                                      & 580                                                         & 580                                                   & 150→300→450→580                                                  \\
Irreversible Path                                                              & Built-In Difficulty                                                                      & .9                                                                                  & .9                                                                 & .1→.3→.5→.7                                                      & .9                                                          & .9                                                    & .1→.3→.5→.7→.9                                                   \\ \bottomrule
\end{tabular}

}
\end{table}

\subsection{Baselines}
The primary goal of these case studies is to assess the effectiveness of our proposed cross-episodic curriculum in increasing the sample efficiency and boosting the generalization capability of Transformer agents. Therefore, in online RL settings, we compare against source RL agents which generate training data for our method and refer to them as \textit{oracles}. These include a) PPO agents directly trained on test task distributions, denoted as ``\textbf{RL (Oracle)}'' hereafter, and b) curriculum PPO agents that are gradually adapted from easier tasks to the test difficulty, which is referred to as ``\textbf{Curriculum RL (Oracle)}''.
Furthermore, we compare against one concurrent and competitive method Agentic Transformer~\citep{liu2023emergent}, denoted as ``\textbf{AT}''. It is closely related to our method, training Transformers on sequences of trajectory ascending sorted according to their rewards.
We also compare against popular offline RL method Decision Transformer~\citep{chen2021decisiontransformer}, denoted as ``\textbf{DT}''.
Additionally, we include another behavior cloning agent that has the same model architecture as ours but is trained on optimal data without cross-episodic attention. 
This baseline is denoted as ``\textbf{BC w/ Expert Data}''.
For the case study on IL from mixed-quality demonstrations, we adopt the most competing approach, \textbf{BC-RNN}, from \citet{mandlekar2021matters} as the main baseline. We also include comparisons against other offline RL methods~\citep{levine2020offline} such as Batch-Constrained Q-learning~(\textbf{BCQ})~\citep{fujimoto2018offpolicy} and Conservative Q-Learning~(\textbf{CQL})~\citep{kumar2020conservative}.

\subsection{Training and Evaluation}
We follow the best practice to train Transformer agents, including adopting AdamW optimizer~\citep{loshchilov2019decoupled}, learning rate warm-up and cosine annealing~\citep{loshchilov17cosinelr}, etc.
Training is performed on NVIDIA V100 GPUs. 
During evaluation, for agents resulting from our method, each run involves several test rollouts to fill the context. We keep hidden states of Transformer-XL~\citep{dai2019transformerxl} propagating across episodes.
We run other baselines and oracles for 100 episodes to estimate their performances.
For our methods on RL settings, we compute the maximum success rate averaged across a sliding window over all test episodes to account for in-context improvement.
The size of the sliding window equals one-quarter of the total test episodes.
These values are averaged over 20 runs to constitute the final reporting metric.
For our methods on the IL setting, since all training data are successful trajectories, we follow \citet{mandlekar2021matters} to report the maximum success rate achieved over the course of training, directly averaged over test episodes.

\section{Experiments}
\label{sec:experiments}
\begin{figure}[!t]
\vspace{-0.1in}
    \centering
    \makebox[\textwidth][c]{\includegraphics[width=1\textwidth]{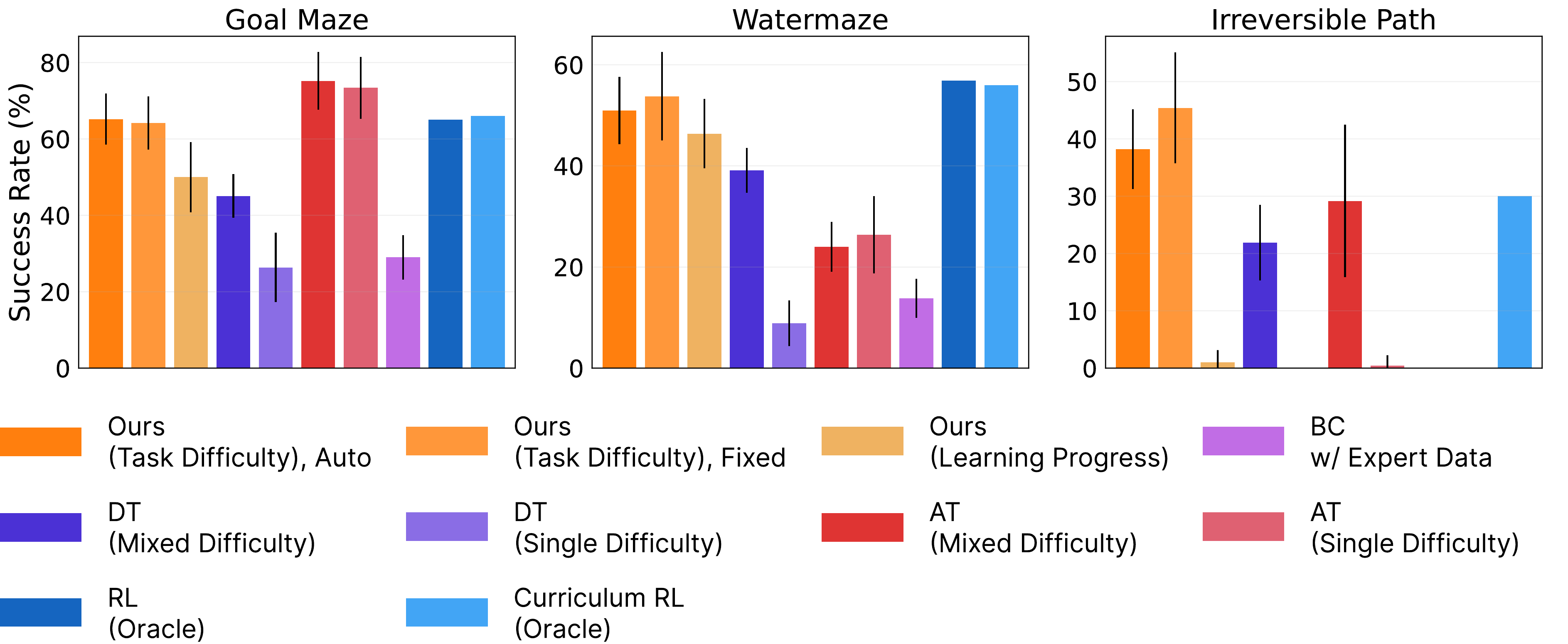}}
    \caption{\textbf{Evaluation results on DMLab.} Our \cec agents perform comparable to RL oracles and on average outperform other baseline methods. On the hardest task Irreversible Path where the RL oracle and BC baseline completely fail, our agents outperform the curriculum RL oracle by $50\%$ even in a \textit{zero-shot} manner. For our methods, DT, AT, and the BC w/ expert data baselines, we conduct 20 independent evaluation runs, each consisting of 100 episodes for Goal Maze and Watermaze and 50 episodes for Irreversible Path due to longer episode length. We test RL oracles for 100 episodes. The error bars represent the standard deviations over 20 runs.}
    \label{fig:dmlab_core}
    \vspace{-1em}
\end{figure}

We aim to answer the following four research questions through comprehensive experiments.

\begin{enumerate}
    \item{To what extent can our cross-episodic curriculum increase the sample efficiency of Transformer agents and boost their generalization capability?}

    \item{Is CEC consistently effective and generally applicable across distinct learning settings?}

    \item{What are the major components that contribute to the effectiveness of our method?}

\end{enumerate}

\vspace{-0.5em}

\subsection{Main Evaluations}
We answer the first two questions above by comparing learned agents from our method against 1) Reinforcement Learning (RL) oracles in online RL settings and 2) well-established baselines on learning from mixed-quality demonstrations in the Imitation Learning (IL) setting.

We first examine agents learned from learning-progress-based and task-difficulty-based curricula in challenging 3D maze environments.
The first type of agent is denoted as ``\textbf{Ours (Learning Progress)}''.
For the second type, to ensure that the evaluation also contains a series of tasks with increasing difficulty, we adopt two mechanisms that control the task sequencing~\citep{Narvekar2017AutonomousTS}: 1) fixed sequencing where agents try each level of difficulty for a fixed amount of times regardless of their performance and 2) dynamic sequencing where agents are automatically promoted to the next difficulty level if they consecutively succeed in the previous level for three times.
We denote these two variants as ``\textbf{Ours (Task Difficulty), Fixed}'' and ``\textbf{Ours (Task Difficulty), Auto}'', respectively.
Note that because the task-difficulty-based curriculum does not contain any training data on test configurations, these two settings are zero-shot evaluated on test task distributions. We summarize these differences in Table~\ref{table:dmlab_exp_setting}. 
We denote AT and DT trained on data consisting of a mixture of task difficulties as ``\textbf{AT (Mixed Difficulty)}'' and ``\textbf{DT (Mixed Difficulty)}''.
Note that these data are the same used to train ``Ours (Task Difficulty)''.
Similarly, we denote AT and DT directly trained on test difficulty as ``AT (Single Difficulty)'' and ``DT (Single Difficulty)''.
These data are the same used to train ``Ours (Learning Progress)''.
\begin{figure}[!t]
    \centering
    \makebox[\textwidth][c]{\includegraphics[width=1\textwidth]{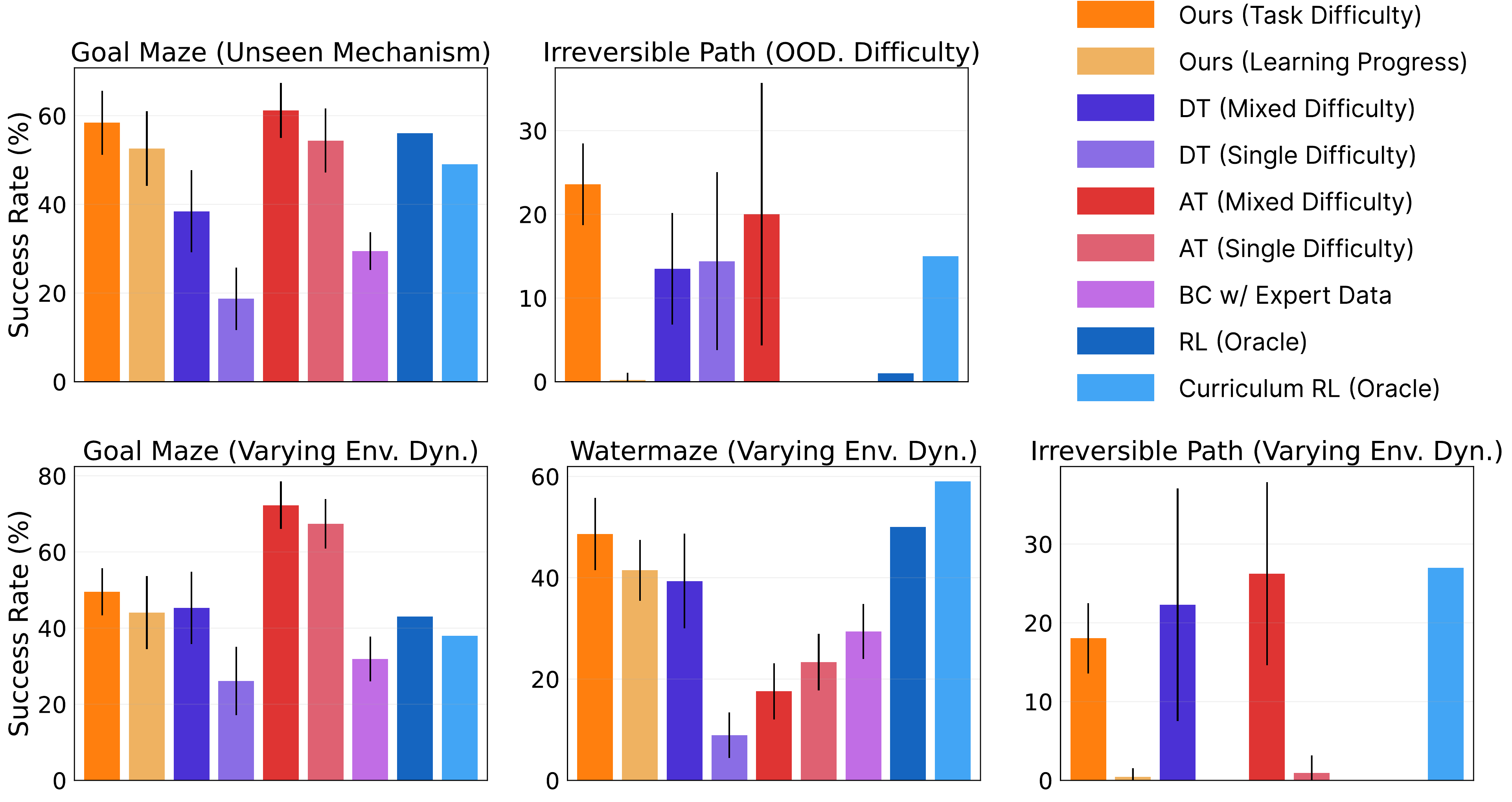}}
    \caption{\textbf{Generalization results on DMLab.} \emph{Top row}: Evaluation results on Goal Maze with unseen maze mechanism and Irreversible Path with out-of-distribution difficulty levels. \emph{Bottom row}: Evaluation results on three levels with environment dynamics differing from training ones. \cec agents display robustness and generalization across various dimensions, outperforming curriculum RL oracles by up to $1.6 \times$. We follow the same evaluation protocol as in Figure~\ref{fig:dmlab_core}. The error bars represent the standard deviations over 20 runs.}
    \label{fig:dmlab_generalization}
    \vspace{-1em}
\end{figure}

\vspace{-0.5em}

\para{Cross-episodic curriculum results in sample-efficient agents.} As shown in Figure~\ref{fig:dmlab_core},  on two out of three examined DMLab levels, \cec agents perform comparable to RL oracles and outperform the BC baselines trained on expert data by at most $2.8\times$.
On the hardest level Irreversible Path where agents have to plan the route ahead and cannot backtrack, both the BC baseline and RL oracle fail.
However, our agents succeed in proposing correct paths that lead to goals and significantly outperform the curriculum RL oracle by $50\%$ even in a \emph{zero-shot} manner.
Because \cec only requires environment interactions generated during the course of training of online source agents (the task-difficulty-based curriculum even contains fewer samples compared to the curriculum RL, as illustrated in Table~\ref{table:dmlab_exp_setting}), the comparable and even better performance demonstrates that our method yields highly sample-efficient embodied policies.
On average, our method with task-difficulty-based curriculum performs the best during evaluation (Table~\ref{supp:table:dmlab_main_avg_results}), confirming the benefit over the concurrent AT approach that leverages chain-of-hindsight experiences. When compared to DT, it outperforms by a significant margin, which suggests that our cross-episodic curriculum helps to squeeze learning signals that are useful for downstream decision-making.

\vspace{-0.5em}

\para{Cross-episodic curriculum boosts the generalization capability.}
To further investigate whether \cec can improve generalization at test time, we construct settings with unseen maze mechanisms (randomly open/closed doors), out-of-distribution difficulty, and different environment dynamics.
See the Appendix, Sec.~\ref{supp:sec:dmlab_generalization} for the exact setups.
As demonstrated in Figure~\ref{fig:dmlab_generalization}, \cec generally improves Transformer agents in learning robust policies that can generalize to perturbations across various axes.
On three settings where the BC w/ Expert Data baseline still manages to make progress, \cec agents are up to $2\times$ better.
Compared to oracle curriculum RL agents, our policies significantly outperform them under three out of five examined scenarios.
It is notable that on Irreversible Path with out-of-distribution difficulty, \cec agent is $1.6\times$ better than the curriculum RL oracle trained on the same data.
These results highlight the benefit of learning with curricular contexts.
On average, our method surpasses the concurrent AT baseline and achieves significantly better performance than other baselines (Table~\ref{supp:table:dmlab_gen_avg_results}). This empirically suggests that CEC helps to learn policies that are robust to environmental perturbations and can quickly generalize to new changes.

\begin{table}[t!]
\caption{\textbf{Evaluation results on RoboMimic.} Visuomotor policies trained with our expertise-based curriculum outperform the most competing history-dependent behavior cloning baseline, as well as other offline RL algorithms. For our method on the Lift task, we conduct 5 independent runs each with 10 rollout episodes. On the Can task, we conduct 10 independent runs each with 5 rollout episodes due to the longer horizon required to complete the task. Standard deviations are included.}
\label{table:robomimic_main}
\vspace{0.1in}
\centering
\resizebox{0.58\textwidth}{!}{
\begin{tabular}{@{}ccccc@{}}
\toprule
\textbf{Task} & \textbf{Ours} & \textbf{BC-RNN}~\citep{mandlekar2021matters}             & \textbf{BCQ}~\citep{fujimoto2018offpolicy}                & \textbf{CQL}~\citep{kumar2020conservative}                \\ \midrule
Lift          &  $\bestscore{100.0 \pm 0.0}$             &    $\bestscore{100.0 \pm 0.0}$                         & $93.3 \pm 0.9$ & $11.3 \pm 9.3$ \\
Can           &   $\bestscore{100.0 \pm 0.0}$            & $\hphantom{\mathbf{0}}96.0 \pm 1.6$ & $77.3 \pm 6.8$ & $\hphantom{0}0.0 \pm 0.0$  \\ \bottomrule
\end{tabular}}

\end{table}

\vspace{-0.5em}

\para{Cross-episodic curriculum is effective across a wide variety of learning scenarios.}
We now move beyond RL settings and study the effectiveness of the expertise-based curriculum in the IL setting with mixed-quality demonstrations.
This is a common scenario, especially in robotics, where demonstrations are collected by human operators with varying proficiency~\citep{mandlekar2018roboturk}.
As presented in Table~\ref{table:robomimic_main}, visuomotor policies trained with the expertise-based curriculum are able to match and outperform the well-established baseline~\citep{mandlekar2021matters} on two simulated robotic manipulation tasks and achieve significantly better performance than agents learned from prevalent offline RL algorithms~\citep{fujimoto2018offpolicy,kumar2020conservative}.
These results suggest that our cross-episodic curriculum is effective and broadly applicable across various problem settings.
More importantly, it provides a promising approach to utilizing limited but sub-optimal data in data-scarce regimes such as robot learning.

\vspace{-0.5em}

\subsection{Ablation Studies}
In this section, we seek to answer the third research question to identify the components critical to the effectiveness of our approach.
We focus on three parts: the importance of cross-episodic attention, the influence of curriculum granularity, and the effect of varying context length.
Finally, we delve into the fourth question, identifying scenarios where CEC is expected to be helpful.

\begin{wrapfigure}{R}{0.4\textwidth}
\vspace{-0.2in}
  \begin{center}
    \includegraphics[width=0.4\textwidth]{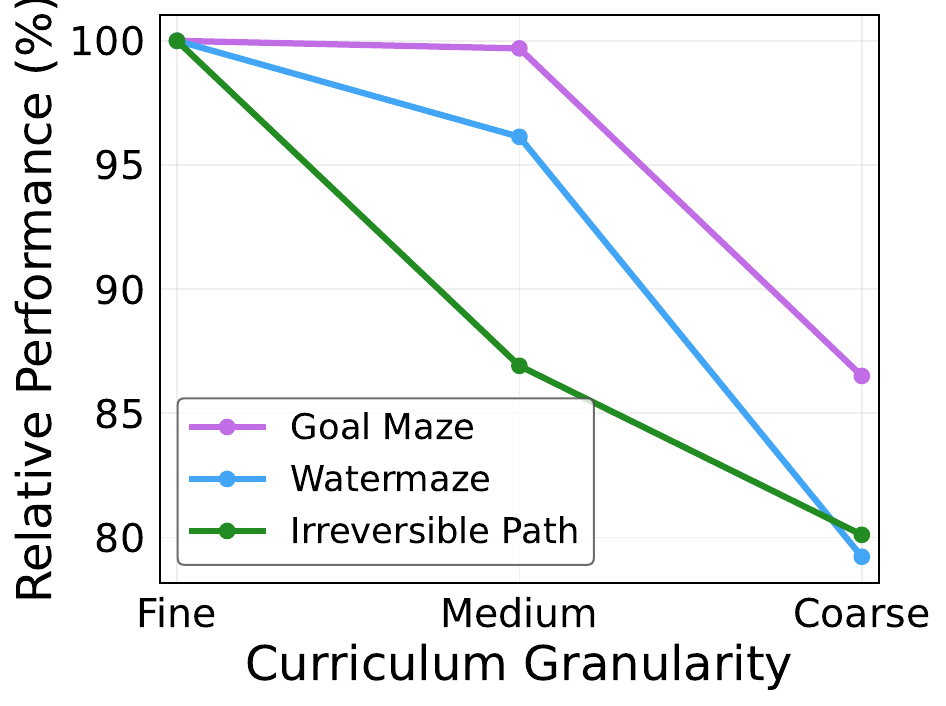}
  \end{center}
      \vspace{-1em}
    \caption{We compare the performance relative to agents trained with the fine-grained curricula. Performance monotonically degrades as task-difficulty-based curricula become coarser.}
    \label{fig:ablation_curriculum_granularity}
\end{wrapfigure}

\begin{table}[t!]
\caption{\textbf{Ablation on the importance of cross-episodic attention.} Transformer agents trained with the same curricular data but without cross-episodic attention degrade significantly during evaluation, suggesting its indispensable role in learning highly performant policies.}
\label{table:ablation_cross_episodic_attention}
\vspace{0.1in}
\centering
\resizebox{1\textwidth}{!}{
\begin{tabular}{@{}r|ccc|cc@{}}
\toprule
\multirow{2}{*}{}                 & \multicolumn{3}{c|}{\textbf{DMLab}}    & \multicolumn{2}{c}{\textbf{RoboMimic}} \\ \cmidrule(l){2-6} 
                                  & Goal Maze   & Watermaze   & Irreversible Path & Lift               & Can               \\ \midrule
Ours                              & $\bestscore{65.2 \pm 6.7}$ & $\bestscore{50.9 \pm 6.6}$ & $\bestscore{38.2 \pm 7.0}$       & $\bestscore{100.0 \pm 0.0}$       & $\bestscore{100.0 \pm 0.0}$      \\
Ours w/o Cross-Episodic Attention & $35.0 \pm 7.1$ & $20.0 \pm 2.5$ & $\hphantom{0}3.8 \pm 4.9$        & $\hphantom{00}75.9 \pm 12.3$       & $\bestscore{\hphantom{0}99.3 \pm 0.9}$       \\ \bottomrule
\end{tabular}

}
\vspace{-1em}
\end{table}

\vspace{-0.5em}

\para{Importance of cross-episodic attention.}
The underlying hypothesis behind our method is that cross-episodic attention enables Transformer agents to distill policy improvement when mixed-optimality trajectories are viewed collectively.
To test this, on DMLab levels and RoboMimic tasks, we train the same Transformer agents with the same curricular data and training epochs but without cross-episodic attention.
We denote such agents as ``\textbf{Ours w/o Cross-Episodic Attention}'' in Table~\ref{table:ablation_cross_episodic_attention}.
Results demonstrate that the ablated variants experience dramatic performance degradation on four out of five examined tasks, which suggests that naively behaviorally cloning sub-optimal data can be problematic and detrimental.
Cross-episodic attention views curricular data collectively, facilitating the extraction of knowledge and patterns crucial for refining decision-making, thereby optimizing the use of sub-optimal data.

\vspace{-0.5em}

\para{Curriculum granularity.} We perform this ablation with the task-difficulty-based curriculum on DMLab levels, due to the ease of adjusting granularity.
We treat the curricula listed in the column ``Ours (Task Difficulty)'' in Table~\ref{table:dmlab_exp_setting} as ``Fine'', and gradually make them coarser to study the impact.
Note that we ensure the same amount of training data.
See the Appendix, Sec.~\ref{supp:sec:curriculum_granularity} for how we define granularity levels ``Medium'' and ``Coarse''.
We visualize the performance relative to the most fine-grained in Figure~\ref{fig:ablation_curriculum_granularity}.
The monotonic degradation of policy performance with respect to curriculum coarseness suggests that fine-grained curricula are critical for Transformer agents to mostly benefit from cross-episodic training.

\begin{wrapfigure}{R}{0.4\textwidth}
\vspace{-2em}
  \begin{center}
    \includegraphics[width=0.4\textwidth]{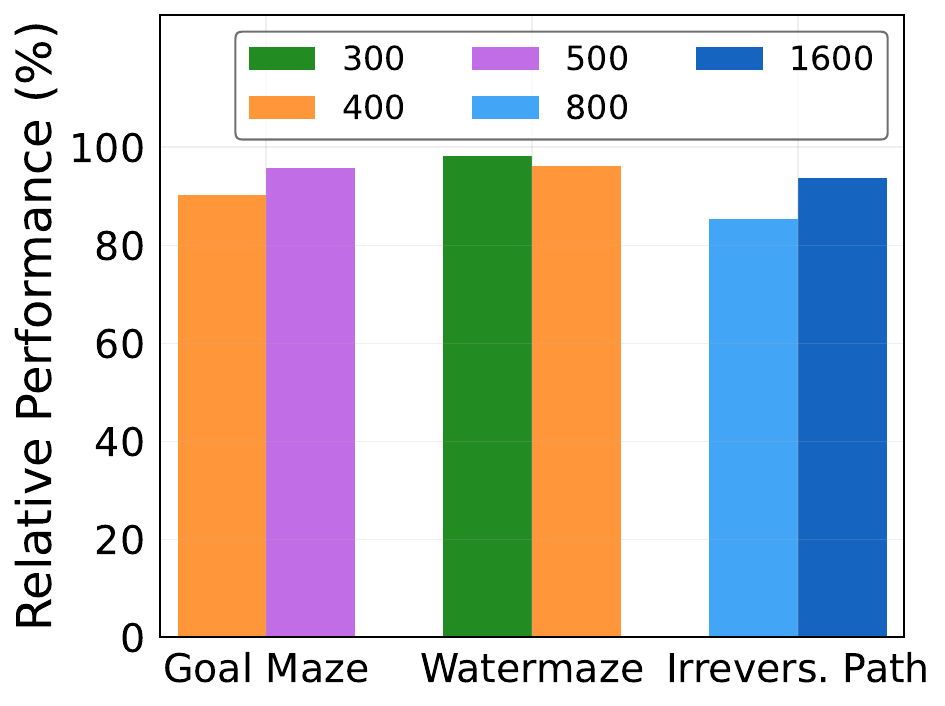}
  \end{center}
        \vspace{-1em}
    \caption{Both short and unnecessarily long context windows decrease the performance. Numbers in the legend denote context lengths. Performance values are relative to those of ``Ours (Task Difficulty), Auto'' reported in Figure~\ref{fig:dmlab_core}. ``Irrevers. Path'' stands for the task ``Irreversible Path''.}
    \label{fig:ablation_horizon}
    \vspace{-3em}
\end{wrapfigure}

\vspace{-0.5em}

\para{Varying context length.}
Lastly, we study the effect of varying context length on DMLab and visualize it in Figure~\ref{fig:ablation_horizon}.
We normalize all performance values relative to those of ``Ours (Task Difficulty), Auto'' reported in Figure~\ref{fig:dmlab_core}.
It turns out that both too short and unnecessarily long context windows are harmful.
On two out of three levels, using a shorter context decreases the performance even more.
This finding coincides with \citet{laskin2023incontext} that a sufficiently long Transformer context is necessary to retain cross-episodic information.
Furthermore, we also discover that an unnecessarily long context is also harmful.
We hypothesize that this is due to the consequent training and optimization instability.

\para{Curriculum selection based on task complexities and data sources.}
For RL tasks, we recommend starting with the learning-progress-based curriculum. However, if the task itself is too challenging, such that source algorithms barely make progress, we recommend the task-difficulty-based curriculum.
In IL settings, we further investigate the performance of the learning-progress-based curriculum on RoboMimic tasks considered in this work. Detailed setup and results are included in Appendix, Sec~\ref{supp:sec:curriculum_comparison}. To summarize, if human demonstrations are available, even if they are generated to be heterogeneous in quality, we recommend using the expertise-based curriculum. However, in the absence of human demonstrations and only with access to machine-generated data (e.g., generated by RL agents), our learning-progress-based curriculum is recommended because it achieves non-trivial performance and significantly outperforms offline RL methods such as CQL~\citep{kumar2020conservative}. 
\section{Related Work}
\label{sec:related_work}

\para{Sequential decision-making with Transformer agents.}
There are many ongoing efforts to replicate the strong emergent properties demonstrated by Transformer models for sequential decision-making problems~\citep{yang2023foundation}.
Decision Transformer~\citep{chen2021decisiontransformer} and Trajectory Transformer~\citep{janner2021onebigsequence} pioneered this thread by casting offline RL~\citep{levine2020offline} as sequence modeling problems.
Gato~\citep{reed2022gato} learns a massively multi-task agent that can be prompted to complete embodied tasks.
MineDojo~\citep{fan2022minedojo} and VPT~\citep{openai2022vpt} utilize numerous YouTube videos for large-scale pre-training in the video game \textit{Minecraft}.
VIMA~\citep{jiang2022vima} and RT-1~\citep{brohan2022rt1} build Transformer agents trained at scale for robotic manipulation tasks.
BeT~\citep{shafiullah2022behavior} and C-BeT~\citep{cui2022play} design novel techniques to learn from demonstrations with multiple modes with Transformers.
Our causal policy most resembles to VPT~\citep{openai2022vpt}. But we focus on designing learning techniques that are generally effective across a wide spectrum of learning scenarios and application domains.

\vspace{-0.5em}

\para{Cross-episodic learning.}
Cross-episodic learning is a less-explored terrain despite that it has been discussed together with meta-RL~\citep{wang2016learning} for a long time.
RL$^2$~\citep{duan2016rl2} uses recurrent neural networks for online meta-RL by optimizing multi-episodic value functions.
Meta-Q-learning~\citep{fakoor2019metaqlearning} instead learns multi-episodic value functions in an offline manner.
Algorithm Distillation (AD)~\citep{laskin2023incontext} and Adaptive Agent (AdA)~\citep{team2023humantimescale} are two recent, inspiring methods in cross-episodic learning. 
Though at first glance our learning-progress-based curriculum appears similar to AD, significant differences emerge. Unlike AD, which focuses on in-context improvements at test time and requires numerous single-task source agents for data generation, our approach improves data efficiency for Transformer agents by structuring data in curricula, requiring only a single multi-task agent and allowing for diverse task instances during evaluations. 
Meanwhile, AdA, although using cross-episodic attention with a Transformer backbone, is rooted in online RL within a proprietary environment. 
In contrast, we focus on offline behavior cloning in accessible, open-source environments, also extending to IL scenarios unexplored by other meta-learning techniques. 
Complementary to this, another recent study~\citep{lee2023supervised} provides theoretical insight into cross-episodic learning.

\vspace{-0.5em}

\para{Curriculum learning.}
Curriculum learning represents training strategies that organize learning samples in meaningful orders to facilitate learning~\citep{bengio2009curriculum}.
It has been proven effective in numerous works that adaptively select simpler task~\citep{Narvekar2017AutonomousTS, Svetlik2017AutomaticCG, riedmiller2018learning, Peng2018CurriculumDF, Czarnecki2018MixMatchA, matiisen2019teacher, narvekar2019learning, Lin2019AdaptiveAT} or auxiliary rewards\citep{Jaderberg2017ReinforcementLW, shen2019situational}. Tasks are also parameterized to form curricula by manipulating goals~\citep{Forestier2017IntrinsicallyMG, Held2018AutomaticGG, Racanire2020AutomatedCG}, environment layouts\citep{wohlke2020performance, baker2019emergent, Portelas2019TeacherAF}, and reward functions~\citep{Gupta2018UnsupervisedMF, Jabri2019UnsupervisedCF}.
Inspired by this paradigm, our work harnesses the improving nature of sequential experiences to boost learning efficiency and generalization for embodied tasks.

\vspace{-0.5em}

\section{Conclusion}
\label{sec:conclusion}
In this work, we introduce a new learning algorithm named \emph{Cross-Episodic Curriculum} to enhance the sample efficiency of policy learning and generalization capability of Transformer agents.
It leverages the shifting distributions of past learning experiences or human demonstrations when they are viewed as curricula.
Combined with cross-episodic attention, \cec yields embodied policies that attain high performance and robust generalization across distinct and representative RL and IL settings.
\cec represents a solid step toward sample-efficient policy learning and is promising for data-scarce problems and real-world domains.

\para{Limitations and future work.}
The CEC algorithm relies on the accurate formulation of curricular sequences that capture the improving nature of multiple experiences. However, defining these sequences accurately can be challenging, especially when dealing with complex environments or tasks. Incorrect or suboptimal formulations of these sequences could negatively impact the algorithm's effectiveness and the overall learning efficiency of the agents.
A thorough exploration regarding the attainability of curricular data is elaborated upon in Appendix, Sec \ref{supp:sec:feasibility}.

In subsequent research, the applicability of \cec to real-world tasks, especially where task difficulty remains ambiguous, merits investigation. A deeper assessment of a demonstrator's proficiency trajectory --- from initial unfamiliarity to the establishment of muscle memory --- could offer a valuable learning signal. Moreover, integrating real-time human feedback to dynamically adjust the curriculum poses an intriguing challenge, potentially enabling \cec to efficiently operate in extended contexts, multi-agent environments, and tangible real-world tasks.

\begin{ack}
We thank Guanzhi Wang and Annie Xie for helpful discussions.
We are grateful to Yifeng Zhu, Zhenyu Jiang, Soroush Nasiriany, Huihan Liu, and Rutav Shah for constructive feedback on an early draft of this paper.
We also thank the anonymous reviewers for offering us insightful suggestions and kind encouragement during the review period.
This work was partially supported by research funds from Salesforce and JP Morgan.
\end{ack}

\bibliographystyle{plainnat}

\newpage
\begin{thebibliography}{82}
\providecommand{\natexlab}[1]{#1}
\providecommand{\url}[1]{\texttt{#1}}
\expandafter\ifx\csname urlstyle\endcsname\relax
  \providecommand{\doi}[1]{doi: #1}\else
  \providecommand{\doi}{doi: \begingroup \urlstyle{rm}\Url}\fi

\bibitem[{Adaptive Agent Team} et~al.(2023){Adaptive Agent Team}, Bauer, Baumli, Baveja, Behbahani, Bhoopchand, Bradley-Schmieg, Chang, Clay, Collister, Dasagi, Gonzalez, Gregor, Hughes, Kashem, Loks-Thompson, Openshaw, Parker-Holder, Pathak, Perez-Nieves, Rakicevic, Rocktäschel, Schroecker, Sygnowski, Tuyls, York, Zacherl, and Zhang]{team2023humantimescale}
{Adaptive Agent Team}, Jakob Bauer, Kate Baumli, Satinder Baveja, Feryal Behbahani, Avishkar Bhoopchand, Nathalie Bradley-Schmieg, Michael Chang, Natalie Clay, Adrian Collister, Vibhavari Dasagi, Lucy Gonzalez, Karol Gregor, Edward Hughes, Sheleem Kashem, Maria Loks-Thompson, Hannah Openshaw, Jack Parker-Holder, Shreya Pathak, Nicolas Perez-Nieves, Nemanja Rakicevic, Tim Rocktäschel, Yannick Schroecker, Jakub Sygnowski, Karl Tuyls, Sarah York, Alexander Zacherl, and Lei Zhang.
\newblock Human-timescale adaptation in an open-ended task space.
\newblock \emph{arXiv preprint arXiv: Arxiv-2301.07608}, 2023.

\bibitem[Ahn et~al.(2022)Ahn, Brohan, Brown, Chebotar, Cortes, David, Finn, Gopalakrishnan, Hausman, Herzog, Ho, Hsu, Ibarz, Ichter, Irpan, Jang, Ruano, Jeffrey, Jesmonth, Joshi, Julian, Kalashnikov, Kuang, Lee, Levine, Lu, Luu, Parada, Pastor, Quiambao, Rao, Rettinghouse, Reyes, Sermanet, Sievers, Tan, Toshev, Vanhoucke, Xia, Xiao, Xu, Xu, and Yan]{ahn2022saycan}
Michael Ahn, Anthony Brohan, Noah Brown, Yevgen Chebotar, Omar Cortes, Byron David, Chelsea Finn, Keerthana Gopalakrishnan, Karol Hausman, Alex Herzog, Daniel Ho, Jasmine Hsu, Julian Ibarz, Brian Ichter, Alex Irpan, Eric Jang, Rosario~Jauregui Ruano, Kyle Jeffrey, Sally Jesmonth, Nikhil~J Joshi, Ryan Julian, Dmitry Kalashnikov, Yuheng Kuang, Kuang-Huei Lee, Sergey Levine, Yao Lu, Linda Luu, Carolina Parada, Peter Pastor, Jornell Quiambao, Kanishka Rao, Jarek Rettinghouse, Diego Reyes, Pierre Sermanet, Nicolas Sievers, Clayton Tan, Alexander Toshev, Vincent Vanhoucke, Fei Xia, Ted Xiao, Peng Xu, Sichun Xu, and Mengyuan Yan.
\newblock Do as i can, not as i say: Grounding language in robotic affordances.
\newblock \emph{arXiv preprint arXiv: Arxiv-2204.01691}, 2022.

\bibitem[Baker et~al.(2020)Baker, Kanitscheider, Markov, Wu, Powell, McGrew, and Mordatch]{baker2019emergent}
Bowen Baker, Ingmar Kanitscheider, Todor Markov, Yi~Wu, Glenn Powell, Bob McGrew, and Igor Mordatch.
\newblock Emergent tool use from multi-agent autocurricula.
\newblock In \emph{International Conference on Learning Representations}, 2020.

\bibitem[Baker et~al.(2022)Baker, Akkaya, Zhokhov, Huizinga, Tang, Ecoffet, Houghton, Sampedro, and Clune]{openai2022vpt}
Bowen Baker, Ilge Akkaya, Peter Zhokhov, Joost Huizinga, Jie Tang, Adrien Ecoffet, Brandon Houghton, Raul Sampedro, and Jeff Clune.
\newblock Video pretraining (vpt): Learning to act by watching unlabeled online videos.
\newblock \emph{arXiv preprint arXiv: Arxiv-2206.11795}, 2022.

\bibitem[Beattie et~al.(2016)Beattie, Leibo, Teplyashin, Ward, Wainwright, Küttler, Lefrancq, Green, Valdés, Sadik, Schrittwieser, Anderson, York, Cant, Cain, Bolton, Gaffney, King, Hassabis, Legg, and Petersen]{beattie2016deepmind}
Charles Beattie, Joel~Z. Leibo, Denis Teplyashin, Tom Ward, Marcus Wainwright, Heinrich Küttler, Andrew Lefrancq, Simon Green, Víctor Valdés, Amir Sadik, Julian Schrittwieser, Keith Anderson, Sarah York, Max Cant, Adam Cain, Adrian Bolton, Stephen Gaffney, Helen King, Demis Hassabis, Shane Legg, and Stig Petersen.
\newblock Deepmind lab.
\newblock \emph{arXiv preprint arXiv: Arxiv-1612.03801}, 2016.

\bibitem[Beliaev et~al.(2022)Beliaev, Shih, Ermon, Sadigh, and Pedarsani]{beliaev2022imitation}
M.~Beliaev, Andy Shih, S.~Ermon, Dorsa Sadigh, and Ramtin Pedarsani.
\newblock Imitation learning by estimating expertise of demonstrators.
\newblock \emph{International Conference On Machine Learning}, 2022.

\bibitem[Bengio et~al.(2009)Bengio, Louradour, Collobert, and Weston]{bengio2009curriculum}
Yoshua Bengio, J{\'e}r{\^o}me Louradour, Ronan Collobert, and Jason Weston.
\newblock Curriculum learning.
\newblock In \emph{International Conference on Machine Learning}, 2009.

\bibitem[Bommasani et~al.(2021)Bommasani, Hudson, Adeli, Altman, Arora, von Arx, Bernstein, Bohg, Bosselut, Brunskill, Brynjolfsson, Buch, Card, Castellon, Chatterji, Chen, Creel, Davis, Demszky, Donahue, Doumbouya, Durmus, Ermon, Etchemendy, Ethayarajh, Fei-Fei, Finn, Gale, Gillespie, Goel, Goodman, Grossman, Guha, Hashimoto, Henderson, Hewitt, Ho, Hong, Hsu, Huang, Icard, Jain, Jurafsky, Kalluri, Karamcheti, Keeling, Khani, Khattab, Koh, Krass, Krishna, Kuditipudi, Kumar, Ladhak, Lee, Lee, Leskovec, Levent, Li, Li, Ma, Malik, Manning, Mirchandani, Mitchell, Munyikwa, Nair, Narayan, Narayanan, Newman, Nie, Niebles, Nilforoshan, Nyarko, Ogut, Orr, Papadimitriou, Park, Piech, Portelance, Potts, Raghunathan, Reich, Ren, Rong, Roohani, Ruiz, Ryan, Ré, Sadigh, Sagawa, Santhanam, Shih, Srinivasan, Tamkin, Taori, Thomas, Tramèr, Wang, Wang, Wu, Wu, Wu, Xie, Yasunaga, You, Zaharia, Zhang, Zhang, Zhang, Zhang, Zheng, Zhou, and Liang]{bommasani2021opportunities}
Rishi Bommasani, Drew~A. Hudson, Ehsan Adeli, Russ Altman, Simran Arora, Sydney von Arx, Michael~S. Bernstein, Jeannette Bohg, Antoine Bosselut, Emma Brunskill, Erik Brynjolfsson, Shyamal Buch, Dallas Card, Rodrigo Castellon, Niladri Chatterji, Annie Chen, Kathleen Creel, Jared~Quincy Davis, Dora Demszky, Chris Donahue, Moussa Doumbouya, Esin Durmus, Stefano Ermon, John Etchemendy, Kawin Ethayarajh, Li~Fei-Fei, Chelsea Finn, Trevor Gale, Lauren Gillespie, Karan Goel, Noah Goodman, Shelby Grossman, Neel Guha, Tatsunori Hashimoto, Peter Henderson, John Hewitt, Daniel~E. Ho, Jenny Hong, Kyle Hsu, Jing Huang, Thomas Icard, Saahil Jain, Dan Jurafsky, Pratyusha Kalluri, Siddharth Karamcheti, Geoff Keeling, Fereshte Khani, Omar Khattab, Pang~Wei Koh, Mark Krass, Ranjay Krishna, Rohith Kuditipudi, Ananya Kumar, Faisal Ladhak, Mina Lee, Tony Lee, Jure Leskovec, Isabelle Levent, Xiang~Lisa Li, Xuechen Li, Tengyu Ma, Ali Malik, Christopher~D. Manning, Suvir Mirchandani, Eric Mitchell, Zanele Munyikwa, Suraj Nair,
  Avanika Narayan, Deepak Narayanan, Ben Newman, Allen Nie, Juan~Carlos Niebles, Hamed Nilforoshan, Julian Nyarko, Giray Ogut, Laurel Orr, Isabel Papadimitriou, Joon~Sung Park, Chris Piech, Eva Portelance, Christopher Potts, Aditi Raghunathan, Rob Reich, Hongyu Ren, Frieda Rong, Yusuf Roohani, Camilo Ruiz, Jack Ryan, Christopher Ré, Dorsa Sadigh, Shiori Sagawa, Keshav Santhanam, Andy Shih, Krishnan Srinivasan, Alex Tamkin, Rohan Taori, Armin~W. Thomas, Florian Tramèr, Rose~E. Wang, William Wang, Bohan Wu, Jiajun Wu, Yuhuai Wu, Sang~Michael Xie, Michihiro Yasunaga, Jiaxuan You, Matei Zaharia, Michael Zhang, Tianyi Zhang, Xikun Zhang, Yuhui Zhang, Lucia Zheng, Kaitlyn Zhou, and Percy Liang.
\newblock On the opportunities and risks of foundation models.
\newblock \emph{arXiv preprint arXiv: Arxiv-2108.07258}, 2021.

\bibitem[Brohan et~al.(2022)Brohan, Brown, Carbajal, Chebotar, Dabis, Finn, Gopalakrishnan, Hausman, Herzog, Hsu, Ibarz, Ichter, Irpan, Jackson, Jesmonth, Joshi, Julian, Kalashnikov, Kuang, Leal, Lee, Levine, Lu, Malla, Manjunath, Mordatch, Nachum, Parada, Peralta, Perez, Pertsch, Quiambao, Rao, Ryoo, Salazar, Sanketi, Sayed, Singh, Sontakke, Stone, Tan, Tran, Vanhoucke, Vega, Vuong, Xia, Xiao, Xu, Xu, Yu, and Zitkovich]{brohan2022rt1}
Anthony Brohan, Noah Brown, Justice Carbajal, Yevgen Chebotar, Joseph Dabis, Chelsea Finn, Keerthana Gopalakrishnan, Karol Hausman, Alex Herzog, Jasmine Hsu, Julian Ibarz, Brian Ichter, Alex Irpan, Tomas Jackson, Sally Jesmonth, Nikhil~J Joshi, Ryan Julian, Dmitry Kalashnikov, Yuheng Kuang, Isabel Leal, Kuang-Huei Lee, Sergey Levine, Yao Lu, Utsav Malla, Deeksha Manjunath, Igor Mordatch, Ofir Nachum, Carolina Parada, Jodilyn Peralta, Emily Perez, Karl Pertsch, Jornell Quiambao, Kanishka Rao, Michael Ryoo, Grecia Salazar, Pannag Sanketi, Kevin Sayed, Jaspiar Singh, Sumedh Sontakke, Austin Stone, Clayton Tan, Huong Tran, Vincent Vanhoucke, Steve Vega, Quan Vuong, Fei Xia, Ted Xiao, Peng Xu, Sichun Xu, Tianhe Yu, and Brianna Zitkovich.
\newblock Rt-1: Robotics transformer for real-world control at scale.
\newblock \emph{arXiv preprint arXiv: Arxiv-2212.06817}, 2022.

\bibitem[Brown et~al.(2019)Brown, Goo, Nagarajan, and Niekum]{brown2019extrapolating}
Daniel~S. Brown, Wonjoon Goo, Prabhat Nagarajan, and Scott Niekum.
\newblock Extrapolating beyond suboptimal demonstrations via inverse reinforcement learning from observations.
\newblock \emph{arXiv preprint arXiv: Arxiv-1904.06387}, 2019.

\bibitem[Cao and Sadigh(2021)]{cao2021learning}
Zhangjie Cao and Dorsa Sadigh.
\newblock Learning from imperfect demonstrations from agents with varying dynamics.
\newblock \emph{arXiv preprint arXiv: Arxiv-2103.05910}, 2021.

\bibitem[Chen et~al.(2020)Chen, Paleja, and Gombolay]{chen2020learning}
Letian Chen, Rohan Paleja, and Matthew Gombolay.
\newblock Learning from suboptimal demonstration via self-supervised reward regression.
\newblock \emph{arXiv preprint arXiv: Arxiv-2010.11723}, 2020.

\bibitem[Chen et~al.(2021)Chen, Lu, Rajeswaran, Lee, Grover, Laskin, Abbeel, Srinivas, and Mordatch]{chen2021decisiontransformer}
Lili Chen, Kevin Lu, Aravind Rajeswaran, Kimin Lee, Aditya Grover, Michael Laskin, Pieter Abbeel, Aravind Srinivas, and Igor Mordatch.
\newblock Decision transformer: Reinforcement learning via sequence modeling.
\newblock In Marc'Aurelio Ranzato, Alina Beygelzimer, Yann~N. Dauphin, Percy Liang, and Jennifer~Wortman Vaughan, editors, \emph{Advances in Neural Information Processing Systems 34: Annual Conference on Neural Information Processing Systems 2021, NeurIPS 2021, December 6-14, 2021, virtual}, pages 15084--15097, 2021.
\newblock URL \url{https://proceedings.neurips.cc/paper/2021/hash/7f489f642a0ddb10272b5c31057f0663-Abstract.html}.

\bibitem[Cui et~al.(2022)Cui, Wang, Shafiullah, and Pinto]{cui2022play}
Zichen~Jeff Cui, Yibin Wang, Nur Muhammad~Mahi Shafiullah, and Lerrel Pinto.
\newblock From play to policy: Conditional behavior generation from uncurated robot data.
\newblock \emph{arXiv preprint arXiv: Arxiv-2210.10047}, 2022.

\bibitem[Czarnecki et~al.(2018)Czarnecki, Jayakumar, Jaderberg, Hasenclever, Teh, Heess, Osindero, and Pascanu]{Czarnecki2018MixMatchA}
Wojciech Czarnecki, Siddhant~M. Jayakumar, Max Jaderberg, Leonard Hasenclever, Yee~Whye Teh, Nicolas Manfred~Otto Heess, Simon Osindero, and Razvan Pascanu.
\newblock Mix\&match - agent curricula for reinforcement learning.
\newblock In \emph{International Conference on Machine Learning}, 2018.

\bibitem[Dai et~al.(2019)Dai, Yang, Yang, Carbonell, Le, and Salakhutdinov]{dai2019transformerxl}
Zihang Dai, Zhilin Yang, Yiming Yang, Jaime Carbonell, Quoc~V. Le, and Ruslan Salakhutdinov.
\newblock Transformer-xl: Attentive language models beyond a fixed-length context.
\newblock \emph{arXiv preprint arXiv: Arxiv-1901.02860}, 2019.

\bibitem[Driess et~al.(2023)Driess, Xia, Sajjadi, Lynch, Chowdhery, Ichter, Wahid, Tompson, Vuong, Yu, Huang, Chebotar, Sermanet, Duckworth, Levine, Vanhoucke, Hausman, Toussaint, Greff, Zeng, Mordatch, and Florence]{driess2023palme}
Danny Driess, Fei Xia, Mehdi S.~M. Sajjadi, Corey Lynch, Aakanksha Chowdhery, Brian Ichter, Ayzaan Wahid, Jonathan Tompson, Quan Vuong, Tianhe Yu, Wenlong Huang, Yevgen Chebotar, Pierre Sermanet, Daniel Duckworth, Sergey Levine, Vincent Vanhoucke, Karol Hausman, Marc Toussaint, Klaus Greff, Andy Zeng, Igor Mordatch, and Pete Florence.
\newblock Palm-e: An embodied multimodal language model.
\newblock \emph{arXiv preprint arXiv: Arxiv-2303.03378}, 2023.

\bibitem[Duan et~al.(2016)Duan, Schulman, Chen, Bartlett, Sutskever, and Abbeel]{duan2016rl2}
Yan Duan, John Schulman, Xi~Chen, Peter~L. Bartlett, Ilya Sutskever, and Pieter Abbeel.
\newblock Rl$^2$: Fast reinforcement learning via slow reinforcement learning.
\newblock \emph{arXiv preprint arXiv: Arxiv-1611.02779}, 2016.

\bibitem[Ebert et~al.(2021)Ebert, Yang, Schmeckpeper, Bucher, Georgakis, Daniilidis, Finn, and Levine]{ebert2021bridge}
Frederik Ebert, Yanlai Yang, Karl Schmeckpeper, Bernadette Bucher, Georgios Georgakis, Kostas Daniilidis, Chelsea Finn, and Sergey Levine.
\newblock Bridge data: Boosting generalization of robotic skills with cross-domain datasets.
\newblock \emph{arXiv preprint arXiv: Arxiv-2109.13396}, 2021.

\bibitem[Espeholt et~al.(2018)Espeholt, Soyer, Munos, Simonyan, Mnih, Ward, Doron, Firoiu, Harley, Dunning, Legg, and Kavukcuoglu]{espeholt2018impala}
Lasse Espeholt, Hubert Soyer, Remi Munos, Karen Simonyan, Volodymir Mnih, Tom Ward, Yotam Doron, Vlad Firoiu, Tim Harley, Iain Dunning, Shane Legg, and Koray Kavukcuoglu.
\newblock Impala: Scalable distributed deep-rl with importance weighted actor-learner architectures.
\newblock \emph{arXiv preprint arXiv: Arxiv-1802.01561}, 2018.

\bibitem[Fakoor et~al.(2019)Fakoor, Chaudhari, Soatto, and Smola]{fakoor2019metaqlearning}
Rasool Fakoor, Pratik Chaudhari, Stefano Soatto, and Alexander~J. Smola.
\newblock Meta-q-learning.
\newblock \emph{arXiv preprint arXiv: Arxiv-1910.00125}, 2019.

\bibitem[Fan et~al.(2022)Fan, Wang, Jiang, Mandlekar, Yang, Zhu, Tang, Huang, Zhu, and Anandkumar]{fan2022minedojo}
Linxi Fan, Guanzhi Wang, Yunfan Jiang, Ajay Mandlekar, Yuncong Yang, Haoyi Zhu, Andrew Tang, De-An Huang, Yuke Zhu, and Anima Anandkumar.
\newblock Minedojo: Building open-ended embodied agents with internet-scale knowledge.
\newblock \emph{arXiv preprint arXiv: Arxiv-2206.08853}, 2022.

\bibitem[Finn et~al.(2015)Finn, Tan, Duan, Darrell, Levine, and Abbeel]{finn2015deep}
Chelsea Finn, Xin~Yu Tan, Yan Duan, Trevor Darrell, Sergey Levine, and Pieter Abbeel.
\newblock Deep spatial autoencoders for visuomotor learning.
\newblock \emph{arXiv preprint arXiv: Arxiv-1509.06113}, 2015.

\bibitem[Forestier et~al.(2017)Forestier, Mollard, and Oudeyer]{Forestier2017IntrinsicallyMG}
S{\'e}bastien Forestier, Yoan Mollard, and Pierre-Yves Oudeyer.
\newblock Intrinsically motivated goal exploration processes with automatic curriculum learning.
\newblock \emph{ArXiv}, abs/1708.02190, 2017.

\bibitem[Fujimoto et~al.(2018)Fujimoto, Meger, and Precup]{fujimoto2018offpolicy}
Scott Fujimoto, David Meger, and Doina Precup.
\newblock Off-policy deep reinforcement learning without exploration.
\newblock \emph{arXiv preprint arXiv: Arxiv-1812.02900}, 2018.

\bibitem[Ghosh et~al.(2019)Ghosh, Gupta, Reddy, Fu, Devin, Eysenbach, and Levine]{ghosh2019learning}
Dibya Ghosh, Abhishek Gupta, Ashwin Reddy, Justin Fu, Coline Devin, Benjamin Eysenbach, and Sergey Levine.
\newblock Learning to reach goals via iterated supervised learning.
\newblock \emph{arXiv preprint arXiv: Arxiv-1912.06088}, 2019.

\bibitem[Graves et~al.(2017)Graves, Bellemare, Menick, Munos, and Kavukcuoglu]{graves2017automated}
Alex Graves, Marc~G. Bellemare, Jacob Menick, Remi Munos, and Koray Kavukcuoglu.
\newblock Automated curriculum learning for neural networks.
\newblock \emph{arXiv preprint arXiv: Arxiv-1704.03003}, 2017.

\bibitem[Gupta et~al.(2018)Gupta, Eysenbach, Finn, and Levine]{Gupta2018UnsupervisedMF}
Abhishek Gupta, Benjamin Eysenbach, Chelsea Finn, and Sergey Levine.
\newblock Unsupervised meta-learning for reinforcement learning.
\newblock \emph{ArXiv}, abs/1806.04640, 2018.

\bibitem[He et~al.(2015)He, Zhang, Ren, and Sun]{he2015resnet}
Kaiming He, Xiangyu Zhang, Shaoqing Ren, and Jian Sun.
\newblock Deep {Residual} {Learning} for {Image} {Recognition}, December 2015.
\newblock URL \url{http://arxiv.org/abs/1512.03385}.
\newblock arXiv:1512.03385 [cs].

\bibitem[Held et~al.(2018)Held, Geng, Florensa, and Abbeel]{Held2018AutomaticGG}
David Held, Xinyang Geng, Carlos Florensa, and Pieter Abbeel.
\newblock Automatic goal generation for reinforcement learning agents.
\newblock In \emph{International Conference on Machine Learning}, 2018.

\bibitem[Hoffman et~al.(2020)Hoffman, Shahriari, Aslanides, Barth-Maron, Momchev, Sinopalnikov, Sta\'nczyk, Ramos, Raichuk, Vincent, Hussenot, Dadashi, Dulac-Arnold, Orsini, Jacq, Ferret, Vieillard, Ghasemipour, Girgin, Pietquin, Behbahani, Norman, Abdolmaleki, Cassirer, Yang, Baumli, Henderson, Friesen, Haroun, Novikov, Colmenarejo, Cabi, Gulcehre, Paine, Srinivasan, Cowie, Wang, Piot, and de~Freitas]{hoffman2020acme}
Matthew~W. Hoffman, Bobak Shahriari, John Aslanides, Gabriel Barth-Maron, Nikola Momchev, Danila Sinopalnikov, Piotr Sta\'nczyk, Sabela Ramos, Anton Raichuk, Damien Vincent, L\'eonard Hussenot, Robert Dadashi, Gabriel Dulac-Arnold, Manu Orsini, Alexis Jacq, Johan Ferret, Nino Vieillard, Seyed Kamyar~Seyed Ghasemipour, Sertan Girgin, Olivier Pietquin, Feryal Behbahani, Tamara Norman, Abbas Abdolmaleki, Albin Cassirer, Fan Yang, Kate Baumli, Sarah Henderson, Abe Friesen, Ruba Haroun, Alex Novikov, Sergio~G\'omez Colmenarejo, Serkan Cabi, Caglar Gulcehre, Tom~Le Paine, Srivatsan Srinivasan, Andrew Cowie, Ziyu Wang, Bilal Piot, and Nando de~Freitas.
\newblock Acme: A research framework for distributed reinforcement learning.
\newblock \emph{arXiv preprint arXiv:2006.00979}, 2020.
\newblock URL \url{https://arxiv.org/abs/2006.00979}.

\bibitem[Huang et~al.(2022{\natexlab{a}})Huang, Abbeel, Pathak, and Mordatch]{huang2022language}
Wenlong Huang, Pieter Abbeel, Deepak Pathak, and Igor Mordatch.
\newblock Language models as zero-shot planners: Extracting actionable knowledge for embodied agents.
\newblock In Kamalika Chaudhuri, Stefanie Jegelka, Le~Song, Csaba Szepesv{\'{a}}ri, Gang Niu, and Sivan Sabato, editors, \emph{International Conference on Machine Learning, {ICML} 2022, 17-23 July 2022, Baltimore, Maryland, {USA}}, volume 162 of \emph{Proceedings of Machine Learning Research}, pages 9118--9147. {PMLR}, 2022{\natexlab{a}}.
\newblock URL \url{https://proceedings.mlr.press/v162/huang22a.html}.

\bibitem[Huang et~al.(2022{\natexlab{b}})Huang, Xia, Xiao, Chan, Liang, Florence, Zeng, Tompson, Mordatch, Chebotar, Sermanet, Brown, Jackson, Luu, Levine, Hausman, and Ichter]{huang2022inner}
Wenlong Huang, Fei Xia, Ted Xiao, Harris Chan, Jacky Liang, Pete Florence, Andy Zeng, Jonathan Tompson, Igor Mordatch, Yevgen Chebotar, Pierre Sermanet, Noah Brown, Tomas Jackson, Linda Luu, Sergey Levine, Karol Hausman, and Brian Ichter.
\newblock Inner monologue: Embodied reasoning through planning with language models.
\newblock \emph{arXiv preprint arXiv: Arxiv-2207.05608}, 2022{\natexlab{b}}.

\bibitem[Jabri et~al.(2019)Jabri, Hsu, Eysenbach, Gupta, Efros, Levine, and Finn]{Jabri2019UnsupervisedCF}
Allan Jabri, Kyle Hsu, Ben Eysenbach, Abhishek Gupta, Alexei~A. Efros, Sergey Levine, and Chelsea Finn.
\newblock Unsupervised curricula for visual meta-reinforcement learning.
\newblock In \emph{Advances in Neural Information Processing Systems}, 2019.

\bibitem[Jaderberg et~al.(2017)Jaderberg, Mnih, Czarnecki, Schaul, Leibo, Silver, and Kavukcuoglu]{Jaderberg2017ReinforcementLW}
Max Jaderberg, Volodymyr Mnih, Wojciech Czarnecki, Tom Schaul, Joel~Z. Leibo, David Silver, and Koray Kavukcuoglu.
\newblock Reinforcement learning with unsupervised auxiliary tasks.
\newblock In \emph{International Conference on Learning Representations}, 2017.

\bibitem[Jaderberg et~al.(2019)Jaderberg, Czarnecki, Dunning, Marris, Lever, Castañeda, Beattie, Rabinowitz, Morcos, Ruderman, Sonnerat, Green, Deason, Leibo, Silver, Hassabis, Kavukcuoglu, and Graepel]{for-the-win}
Max Jaderberg, Wojciech~M. Czarnecki, Iain Dunning, Luke Marris, Guy Lever, Antonio~Garcia Castañeda, Charles Beattie, Neil~C. Rabinowitz, Ari~S. Morcos, Avraham Ruderman, Nicolas Sonnerat, Tim Green, Louise Deason, Joel~Z. Leibo, David Silver, Demis Hassabis, Koray Kavukcuoglu, and Thore Graepel.
\newblock Human-level performance in 3d multiplayer games with population-based reinforcement learning.
\newblock \emph{Science}, 364\penalty0 (6443):\penalty0 859--865, 2019.
\newblock \doi{10.1126/science.aau6249}.
\newblock URL \url{https://www.science.org/doi/abs/10.1126/science.aau6249}.

\bibitem[Janner et~al.(2021)Janner, Li, and Levine]{janner2021onebigsequence}
Michael Janner, Qiyang Li, and Sergey Levine.
\newblock Offline reinforcement learning as one big sequence modeling problem.
\newblock In Marc'Aurelio Ranzato, Alina Beygelzimer, Yann~N. Dauphin, Percy Liang, and Jennifer~Wortman Vaughan, editors, \emph{Advances in Neural Information Processing Systems 34: Annual Conference on Neural Information Processing Systems 2021, NeurIPS 2021, December 6-14, 2021, virtual}, pages 1273--1286, 2021.
\newblock URL \url{https://proceedings.neurips.cc/paper/2021/hash/099fe6b0b444c23836c4a5d07346082b-Abstract.html}.

\bibitem[Jiang et~al.(2022)Jiang, Gupta, Zhang, Wang, Dou, Chen, Fei-Fei, Anandkumar, Zhu, and Fan]{jiang2022vima}
Yunfan Jiang, Agrim Gupta, Zichen Zhang, Guanzhi Wang, Yongqiang Dou, Yanjun Chen, Li~Fei-Fei, Anima Anandkumar, Yuke Zhu, and Linxi Fan.
\newblock Vima: General robot manipulation with multimodal prompts.
\newblock \emph{arXiv preprint arXiv: Arxiv-2210.03094}, 2022.

\bibitem[Kaelbling(1993)]{Kaelbling1993LearningTA}
Leslie~Pack Kaelbling.
\newblock Learning to achieve goals.
\newblock In \emph{International Joint Conference on Artificial Intelligence}, 1993.

\bibitem[Kanitscheider et~al.(2021)Kanitscheider, Huizinga, Farhi, Guss, Houghton, Sampedro, Zhokhov, Baker, Ecoffet, Tang, Klimov, and Clune]{kanitscheider2021multitask}
Ingmar Kanitscheider, Joost Huizinga, David Farhi, William~Hebgen Guss, Brandon Houghton, Raul Sampedro, Peter Zhokhov, Bowen Baker, Adrien Ecoffet, Jie Tang, Oleg Klimov, and Jeff Clune.
\newblock Multi-task curriculum learning in a complex, visual, hard-exploration domain: Minecraft.
\newblock \emph{arXiv preprint arXiv: Arxiv-2106.14876}, 2021.

\bibitem[Kumar et~al.(2020)Kumar, Zhou, Tucker, and Levine]{kumar2020conservative}
Aviral Kumar, Aurick Zhou, George Tucker, and Sergey Levine.
\newblock Conservative q-learning for offline reinforcement learning.
\newblock \emph{arXiv preprint arXiv: Arxiv-2006.04779}, 2020.

\bibitem[Laskin et~al.(2023)Laskin, Wang, Oh, Parisotto, Spencer, Steigerwald, Strouse, Hansen, Filos, Brooks, maxime gazeau, Sahni, Singh, and Mnih]{laskin2023incontext}
Michael Laskin, Luyu Wang, Junhyuk Oh, Emilio Parisotto, Stephen Spencer, Richie Steigerwald, DJ~Strouse, Steven~Stenberg Hansen, Angelos Filos, Ethan Brooks, maxime gazeau, Himanshu Sahni, Satinder Singh, and Volodymyr Mnih.
\newblock In-context reinforcement learning with algorithm distillation.
\newblock In \emph{The Eleventh International Conference on Learning Representations}, 2023.
\newblock URL \url{https://openreview.net/forum?id=hy0a5MMPUv}.

\bibitem[Lee et~al.(2023)Lee, Xie, Pacchiano, Chandak, Finn, Nachum, and Brunskill]{lee2023supervised}
Jonathan~N. Lee, Annie Xie, Aldo Pacchiano, Yash Chandak, Chelsea Finn, Ofir Nachum, and Emma Brunskill.
\newblock Supervised pretraining can learn in-context reinforcement learning.
\newblock \emph{arXiv preprint arXiv: Arxiv-2306.14892}, 2023.

\bibitem[Levine et~al.(2020)Levine, Kumar, Tucker, and Fu]{levine2020offline}
Sergey Levine, Aviral Kumar, George Tucker, and Justin Fu.
\newblock Offline reinforcement learning: Tutorial, review, and perspectives on open problems.
\newblock \emph{arXiv preprint arXiv: Arxiv-2005.01643}, 2020.

\bibitem[Liang et~al.(2022)Liang, Huang, Xia, Xu, Hausman, Ichter, Florence, and Zeng]{liang2022code}
Jacky Liang, Wenlong Huang, Fei Xia, Peng Xu, Karol Hausman, Brian Ichter, Pete Florence, and Andy Zeng.
\newblock Code as policies: Language model programs for embodied control.
\newblock \emph{arXiv preprint arXiv: Arxiv-2209.07753}, 2022.

\bibitem[Lin et~al.(2019)Lin, Baweja, Kantor, and Held]{Lin2019AdaptiveAT}
Xingyu Lin, Harjatin~Singh Baweja, George Kantor, and David Held.
\newblock Adaptive auxiliary task weighting for reinforcement learning.
\newblock In \emph{Advances in Neural Information Processing Systems}, 2019.

\bibitem[Liu and Abbeel(2023)]{liu2023emergent}
Hao Liu and Pieter Abbeel.
\newblock Emergent agentic transformer from chain of hindsight experience.
\newblock \emph{arXiv preprint arXiv: Arxiv-2305.16554}, 2023.

\bibitem[Loshchilov and Hutter(2017)]{loshchilov17cosinelr}
Ilya Loshchilov and Frank Hutter.
\newblock {SGDR:} stochastic gradient descent with warm restarts.
\newblock In \emph{5th International Conference on Learning Representations, {ICLR} 2017, Toulon, France, April 24-26, 2017, Conference Track Proceedings}. OpenReview.net, 2017.
\newblock URL \url{https://openreview.net/forum?id=Skq89Scxx}.

\bibitem[Loshchilov and Hutter(2019)]{loshchilov2019decoupled}
Ilya Loshchilov and Frank Hutter.
\newblock Decoupled weight decay regularization.
\newblock In \emph{7th International Conference on Learning Representations, {ICLR} 2019, New Orleans, LA, USA, May 6-9, 2019}. OpenReview.net, 2019.
\newblock URL \url{https://openreview.net/forum?id=Bkg6RiCqY7}.

\bibitem[Ma et~al.(2022)Ma, Sodhani, Jayaraman, Bastani, Kumar, and Zhang]{ma2022vip}
Yecheng~Jason Ma, Shagun Sodhani, Dinesh Jayaraman, Osbert Bastani, Vikash Kumar, and Amy Zhang.
\newblock Vip: Towards universal visual reward and representation via value-implicit pre-training.
\newblock \emph{arXiv preprint arXiv: Arxiv-2210.00030}, 2022.

\bibitem[Majumdar et~al.(2023)Majumdar, Yadav, Arnaud, Ma, Chen, Silwal, Jain, Berges, Abbeel, Malik, Batra, Lin, Maksymets, Rajeswaran, and Meier]{majumdar2023search}
Arjun Majumdar, Karmesh Yadav, Sergio Arnaud, Yecheng~Jason Ma, Claire Chen, Sneha Silwal, Aryan Jain, Vincent-Pierre Berges, Pieter Abbeel, Jitendra Malik, Dhruv Batra, Yixin Lin, Oleksandr Maksymets, Aravind Rajeswaran, and Franziska Meier.
\newblock Where are we in the search for an artificial visual cortex for embodied intelligence?
\newblock \emph{arXiv preprint arXiv: Arxiv-2303.18240}, 2023.

\bibitem[Mandlekar et~al.(2018)Mandlekar, Zhu, Garg, Booher, Spero, Tung, Gao, Emmons, Gupta, Orbay, Savarese, and Fei-Fei]{mandlekar2018roboturk}
Ajay Mandlekar, Yuke Zhu, Animesh Garg, Jonathan Booher, Max Spero, Albert Tung, Julian Gao, John Emmons, Anchit Gupta, Emre Orbay, Silvio Savarese, and Li~Fei-Fei.
\newblock Roboturk: A crowdsourcing platform for robotic skill learning through imitation.
\newblock \emph{arXiv preprint arXiv: Arxiv-1811.02790}, 2018.

\bibitem[Mandlekar et~al.(2021)Mandlekar, Xu, Wong, Nasiriany, Wang, Kulkarni, Fei-Fei, Savarese, Zhu, and Martín-Martín]{mandlekar2021matters}
Ajay Mandlekar, Danfei Xu, Josiah Wong, Soroush Nasiriany, Chen Wang, Rohun Kulkarni, Li~Fei-Fei, Silvio Savarese, Yuke Zhu, and Roberto Martín-Martín.
\newblock What matters in learning from offline human demonstrations for robot manipulation.
\newblock \emph{arXiv preprint arXiv: Arxiv-2108.03298}, 2021.

\bibitem[Matiisen et~al.(2017)Matiisen, Oliver, Cohen, and Schulman]{matiisen2017teacherstudent}
Tambet Matiisen, Avital Oliver, Taco Cohen, and John Schulman.
\newblock Teacher-student curriculum learning.
\newblock \emph{arXiv preprint arXiv: Arxiv-1707.00183}, 2017.

\bibitem[Matiisen et~al.(2019)Matiisen, Oliver, Cohen, and Schulman]{matiisen2019teacher}
Tambet Matiisen, Avital Oliver, Taco Cohen, and John Schulman.
\newblock Teacher-student curriculum learning.
\newblock In \emph{IEEE transactions on neural networks and learning systems}, 2019.

\bibitem[Morris(1981)]{watermaze}
Richard~G.M. Morris.
\newblock Spatial localization does not require the presence of local cues.
\newblock \emph{Learning and Motivation}, 12\penalty0 (2):\penalty0 239--260, 1981.
\newblock \doi{10.1016/0023-9690(81)90020-5}.
\newblock URL \url{https://app.dimensions.ai/details/publication/pub.1028012961}.

\bibitem[Nair et~al.(2022)Nair, Rajeswaran, Kumar, Finn, and Gupta]{nair2022r3m}
Suraj Nair, Aravind Rajeswaran, Vikash Kumar, Chelsea Finn, and Abhinav Gupta.
\newblock R3m: A universal visual representation for robot manipulation.
\newblock \emph{arXiv preprint arXiv: Arxiv-2203.12601}, 2022.

\bibitem[Narvekar et~al.(2017)Narvekar, Sinapov, and Stone]{Narvekar2017AutonomousTS}
S.~Narvekar, J.~Sinapov, and P.~Stone.
\newblock Autonomous task sequencing for customized curriculum design in reinforcement learning.
\newblock In \emph{International Joint Conference on Artificial Intelligence}, 2017.

\bibitem[Narvekar and Stone(2019)]{narvekar2019learning}
Sanmit Narvekar and Peter Stone.
\newblock Learning curriculum policies for reinforcement learning.
\newblock In \emph{Proceedings of the 18th International Conference on Autonomous Agents and MultiAgent Systems}. International Foundation for Autonomous Agents and Multiagent Systems, 2019.

\bibitem[OpenAI et~al.(2019)OpenAI, Berner, Brockman, Chan, Cheung, Debiak, Dennison, Farhi, Fischer, Hashme, Hesse, Józefowicz, Gray, Olsson, Pachocki, Petrov, d.~O.~Pinto, Raiman, Salimans, Schlatter, Schneider, Sidor, Sutskever, Tang, Wolski, and Zhang]{openai2019dota}
OpenAI, Christopher Berner, Greg Brockman, Brooke Chan, Vicki Cheung, Przemyslaw Debiak, Christy Dennison, David Farhi, Quirin Fischer, Shariq Hashme, Chris Hesse, Rafal Józefowicz, Scott Gray, Catherine Olsson, Jakub Pachocki, Michael Petrov, Henrique~P. d.~O.~Pinto, Jonathan Raiman, Tim Salimans, Jeremy Schlatter, Jonas Schneider, Szymon Sidor, Ilya Sutskever, Jie Tang, Filip Wolski, and Susan Zhang.
\newblock Dota 2 with large scale deep reinforcement learning.
\newblock \emph{arXiv preprint arXiv: Arxiv-1912.06680}, 2019.

\bibitem[Paszke et~al.(2019)Paszke, Gross, Massa, Lerer, Bradbury, Chanan, Killeen, Lin, Gimelshein, Antiga, Desmaison, Kopf, Yang, DeVito, Raison, Tejani, Chilamkurthy, Steiner, Fang, Bai, and Chintala]{Paszke2019PyTorch}
Adam Paszke, Sam Gross, Francisco Massa, Adam Lerer, James Bradbury, Gregory Chanan, Trevor Killeen, Zeming Lin, Natalia Gimelshein, Luca Antiga, Alban Desmaison, Andreas Kopf, Edward Yang, Zachary DeVito, Martin Raison, Alykhan Tejani, Sasank Chilamkurthy, Benoit Steiner, Lu~Fang, Junjie Bai, and Soumith Chintala.
\newblock Pytorch: An imperative style, high-performance deep learning library.
\newblock In H.~Wallach, H.~Larochelle, A.~Beygelzimer, F.~d\textquotesingle Alch\'{e}-Buc, E.~Fox, and R.~Garnett, editors, \emph{Advances in Neural Information Processing Systems 32}, pages 8024--8035. Curran Associates, Inc., 2019.

\bibitem[Peng et~al.(2018)Peng, MacGlashan, Loftin, Littman, Roberts, and Taylor]{Peng2018CurriculumDF}
B.~Peng, J.~MacGlashan, R.~Loftin, M.~Littman, D.~Roberts, and Matthew~E. Taylor.
\newblock Curriculum design for machine learners in sequential decision tasks.
\newblock \emph{IEEE Transactions on Emerging Topics in Computational Intelligence}, 2:\penalty0 268--277, 2018.

\bibitem[Petrenko et~al.(2020)Petrenko, Huang, Kumar, Sukhatme, and Koltun]{petrenko2020sample}
Aleksei Petrenko, Zhehui Huang, Tushar Kumar, Gaurav Sukhatme, and Vladlen Koltun.
\newblock Sample factory: Egocentric 3d control from pixels at 100000 fps with asynchronous reinforcement learning.
\newblock \emph{arXiv preprint arXiv: Arxiv-2006.11751}, 2020.

\bibitem[Portelas et~al.(2019{\natexlab{a}})Portelas, Colas, Hofmann, and Oudeyer]{Portelas2019TeacherAF}
R{\'e}my Portelas, C{\'e}dric Colas, Katja Hofmann, and Pierre-Yves Oudeyer.
\newblock Teacher algorithms for curriculum learning of deep rl in continuously parameterized environments.
\newblock In \emph{Conference on Robot Learning}, 2019{\natexlab{a}}.

\bibitem[Portelas et~al.(2019{\natexlab{b}})Portelas, Colas, Hofmann, and Oudeyer]{portelas2019teacher}
Rémy Portelas, Cédric Colas, Katja Hofmann, and Pierre-Yves Oudeyer.
\newblock Teacher algorithms for curriculum learning of deep rl in continuously parameterized environments.
\newblock \emph{arXiv preprint arXiv: Arxiv-1910.07224}, 2019{\natexlab{b}}.

\bibitem[Racani{\`e}re et~al.(2020)Racani{\`e}re, Lampinen, Santoro, Reichert, Firoiu, and Lillicrap]{Racanire2020AutomatedCG}
S{\'e}bastien Racani{\`e}re, Andrew~Kyle Lampinen, Adam Santoro, David~P. Reichert, Vlad Firoiu, and Timothy~P. Lillicrap.
\newblock Automated curriculum generation through setter-solver interactions.
\newblock In \emph{International Conference on Learning Representations}, 2020.

\bibitem[Radosavovic et~al.(2022)Radosavovic, Xiao, James, Abbeel, Malik, and Darrell]{radosavovic2022realworld}
Ilija Radosavovic, Tete Xiao, Stephen James, Pieter Abbeel, Jitendra Malik, and Trevor Darrell.
\newblock Real-world robot learning with masked visual pre-training.
\newblock \emph{arXiv preprint arXiv: Arxiv-2210.03109}, 2022.

\bibitem[Reed et~al.(2022)Reed, Zolna, Parisotto, Colmenarejo, Novikov, Barth-Maron, Gimenez, Sulsky, Kay, Springenberg, Eccles, Bruce, Razavi, Edwards, Heess, Chen, Hadsell, Vinyals, Bordbar, and de~Freitas]{reed2022gato}
Scott Reed, Konrad Zolna, Emilio Parisotto, Sergio~Gomez Colmenarejo, Alexander Novikov, Gabriel Barth-Maron, Mai Gimenez, Yury Sulsky, Jackie Kay, Jost~Tobias Springenberg, Tom Eccles, Jake Bruce, Ali Razavi, Ashley Edwards, Nicolas Heess, Yutian Chen, Raia Hadsell, Oriol Vinyals, Mahyar Bordbar, and Nando de~Freitas.
\newblock A generalist agent.
\newblock \emph{arXiv preprint arXiv: Arxiv-2205.06175}, 2022.

\bibitem[Riedmiller et~al.(2018)Riedmiller, Hafner, Lampe, Neunert, Degrave, Van~de Wiele, Mnih, Heess, and Springenberg]{riedmiller2018learning}
Martin Riedmiller, Roland Hafner, Thomas Lampe, Michael Neunert, Jonas Degrave, Tom Van~de Wiele, Volodymyr Mnih, Nicolas Heess, and Jost~Tobias Springenberg.
\newblock Learning by playing-solving sparse reward tasks from scratch.
\newblock In \emph{International Conference on Machine Learning}, 2018.

\bibitem[Schulman et~al.(2017)Schulman, Wolski, Dhariwal, Radford, and Klimov]{schulman2017proximal}
John Schulman, Filip Wolski, Prafulla Dhariwal, Alec Radford, and Oleg Klimov.
\newblock Proximal policy optimization algorithms.
\newblock \emph{arXiv preprint arXiv: Arxiv-1707.06347}, 2017.

\bibitem[Shafiullah et~al.(2022)Shafiullah, Cui, Altanzaya, and Pinto]{shafiullah2022behavior}
Nur Muhammad~Mahi Shafiullah, Zichen~Jeff Cui, Ariuntuya Altanzaya, and Lerrel Pinto.
\newblock Behavior transformers: Cloning $k$ modes with one stone.
\newblock \emph{arXiv preprint arXiv: Arxiv-2206.11251}, 2022.

\bibitem[Shen et~al.(2019)Shen, Xu, Zhu, Guibas, Fei-Fei, and Savarese]{shen2019situational}
William~B Shen, Danfei Xu, Yuke Zhu, Leonidas~J Guibas, Li~Fei-Fei, and Silvio Savarese.
\newblock Situational fusion of visual representation for visual navigation.
\newblock In \emph{Proceedings of the IEEE International Conference on Computer Vision}, pages 2881--2890, 2019.

\bibitem[Sutton and Barto(2018)]{sutton2018reinforcement}
Richard~S Sutton and Andrew~G Barto.
\newblock \emph{Reinforcement learning: An introduction}.
\newblock MIT press, 2018.

\bibitem[Svetlik et~al.(2017)Svetlik, Leonetti, Sinapov, Shah, Walker, and Stone]{Svetlik2017AutomaticCG}
M.~Svetlik, Matteo Leonetti, J.~Sinapov, Rishi Shah, Nick Walker, and P.~Stone.
\newblock Automatic curriculum graph generation for reinforcement learning agents.
\newblock In \emph{AAAI Conference on Artificial Intelligence}, 2017.

\bibitem[Vaswani et~al.(2017)Vaswani, Shazeer, Parmar, Uszkoreit, Jones, Gomez, Kaiser, and Polosukhin]{vaswani2017attention}
Ashish Vaswani, Noam Shazeer, Niki Parmar, Jakob Uszkoreit, Llion Jones, Aidan~N. Gomez, Lukasz Kaiser, and Illia Polosukhin.
\newblock Attention is all you need.
\newblock \emph{arXiv preprint arXiv: Arxiv-1706.03762}, 2017.

\bibitem[Wang et~al.(2023{\natexlab{a}})Wang, Xie, Jiang, Mandlekar, Xiao, Zhu, Fan, and Anandkumar]{wang2023voyager}
Guanzhi Wang, Yuqi Xie, Yunfan Jiang, Ajay Mandlekar, Chaowei Xiao, Yuke Zhu, Linxi Fan, and Anima Anandkumar.
\newblock Voyager: An open-ended embodied agent with large language models.
\newblock \emph{arXiv preprint arXiv: Arxiv-2305.16291}, 2023{\natexlab{a}}.

\bibitem[Wang et~al.(2016)Wang, Kurth-Nelson, Tirumala, Soyer, Leibo, Munos, Blundell, Kumaran, and Botvinick]{wang2016learning}
Jane~X Wang, Zeb Kurth-Nelson, Dhruva Tirumala, Hubert Soyer, Joel~Z Leibo, Remi Munos, Charles Blundell, Dharshan Kumaran, and Matt Botvinick.
\newblock Learning to reinforcement learn.
\newblock \emph{arXiv preprint arXiv: Arxiv-1611.05763}, 2016.

\bibitem[Wang et~al.(2023{\natexlab{b}})Wang, Cai, Liu, Ma, and Liang]{wang2023describe}
Zihao Wang, Shaofei Cai, Anji Liu, Xiaojian Ma, and Yitao Liang.
\newblock Describe, explain, plan and select: Interactive planning with large language models enables open-world multi-task agents.
\newblock \emph{arXiv preprint arXiv: Arxiv-2302.01560}, 2023{\natexlab{b}}.

\bibitem[W{\"o}hlke et~al.(2020)W{\"o}hlke, Schmitt, and van Hoof]{wohlke2020performance}
Jan W{\"o}hlke, Felix Schmitt, and Herke van Hoof.
\newblock A performance-based start state curriculum framework for reinforcement learning.
\newblock In \emph{Proceedings of the 19th International Conference on Autonomous Agents and MultiAgent Systems}, pages 1503--1511, 2020.

\bibitem[Yang et~al.(2023)Yang, Nachum, Du, Wei, Abbeel, and Schuurmans]{yang2023foundation}
Sherry Yang, Ofir Nachum, Yilun Du, Jason Wei, Pieter Abbeel, and Dale Schuurmans.
\newblock Foundation models for decision making: Problems, methods, and opportunities.
\newblock \emph{arXiv preprint arXiv: Arxiv-2303.04129}, 2023.

\bibitem[Zhang et~al.(2021)Zhang, CAO, Sadigh, and Sui]{NEURIPS2021_670e8a43}
Songyuan Zhang, ZHANGJIE CAO, Dorsa Sadigh, and Yanan Sui.
\newblock Confidence-aware imitation learning from demonstrations with varying optimality.
\newblock In M.~Ranzato, A.~Beygelzimer, Y.~Dauphin, P.S. Liang, and J.~Wortman Vaughan, editors, \emph{Advances in Neural Information Processing Systems}, volume~34, pages 12340--12350. Curran Associates, Inc., 2021.
\newblock URL \url{https://proceedings.neurips.cc/paper_files/paper/2021/file/670e8a43b246801ca1eaca97b3e19189-Paper.pdf}.

\bibitem[Zhu et~al.(2022)Zhu, Joshi, Stone, and Zhu]{zhu2022viola}
Yifeng Zhu, Abhishek Joshi, Peter Stone, and Yuke Zhu.
\newblock Viola: Imitation learning for vision-based manipulation with object proposal priors.
\newblock \emph{arXiv preprint arXiv: Arxiv-2210.11339}, 2022.

\end{thebibliography}
\newpage
\appendix
\renewcommand{\thefigure}{A.\arabic{figure}}
\setcounter{figure}{0}

\renewcommand{\thetable}{A.\arabic{table}}
\setcounter{table}{0}

\section{Model Architecture}
\label{supp:sec:model_architecture}

In this section, we provide comprehensive details about the Transformer model architectures considered in this work.
We implement all models in PyTorch~\citep{Paszke2019PyTorch} and adapt the implementation of Transformer-XL from VPT~\citep{openai2022vpt}.

\subsection{Observation Encoding}
Experiments conducted on both DMLab and RoboMimic include RGB image observations.
For models trained on DMLab, we use a ConvNet~\citep{he2015resnet} similar to the one used in \citet{espeholt2018impala}. 
For models trained on RoboMimic, we follow \citet{mandlekar2021matters} to use a ResNet-18 network~\citep{he2015resnet} followed by a spatial-softmax layer~\citep{finn2015deep}. We use independent and separate encoders for images taken from the wrist camera and frontal camera.
Detailed model parameters are listed in Table~\ref{supp:table:vision_encoder}.

\begin{table}[h]
\caption{Model hyperparameters for vision encoders.}
\vspace{0.1in}
\centering
\begin{tabular}{@{}ll@{}}
\toprule
\textbf{Hyperparameter} & \textbf{Value}    \\ \midrule
\multicolumn{2}{c}{DMLab} \\\midrule
Image Size     & 72 $\times$ 96 \\
Number of ConvNet Blocks     & 1       \\
Channels per Block      & [16, 32, 32]      \\
Output Size     & 256        \\\midrule
\multicolumn{2}{c}{RoboMimic} \\\midrule
Image Size     & 84 $\times$ 84 \\
Random Crop Height     & 76       \\
Random Crop Width     & 76       \\
Number of Randomly Cropped Patches     & 1       \\
ConvNet Backbone      & ResNet-18~\citep{he2015resnet}      \\
Output Size     & 64        \\
Spatial-Softmax Number of Keypoints     & 32        \\
Spatial-Softmax Temperature     & 1.0        \\
Output Size     & 64        \\\bottomrule
\end{tabular}
\label{supp:table:vision_encoder}
\end{table}

Since DMLab is highly partially observable, we follow previous work~\citep{espeholt2018impala,fan2022minedojo,openai2022vpt} to supply the model with previous action input.
We learn 16-dim embedding vectors for all discrete actions.

To encode proprioceptive measurement in RoboMimic, we follow \citet{mandlekar2021matters} to not apply any learned encoding. Instead, these types of observation are concatenated with image features and passed altogether to the following layers.
Note that we do not provide previous action inputs in RoboMimic, since we find doing so would incur significant overfitting.

\subsection{Transformer Backbone}
We use Transformer-XL~\citep{dai2019transformerxl} as our model backbone, adapted from \citet{openai2022vpt}. Transformer-XL splits long sequences into shorter sub-sequences that reduce the computational cost of attention while allowing the hidden states to be carried across the entire input by attending to previous keys and values.
This feature is critical for the long sequence inputs necessary for cross-episodic attention.
Detailed model parameters are listed in Table~\ref{supp:table:xf_xl}.

\begin{table}[ht]
\caption{Model hyperparameters for Transformer-XL.}
\vspace{0.1in}
\centering
\begin{tabular}{@{}lll@{}}
\toprule
\textbf{Hyperparameter} & \textbf{Value (DMLab)} & \textbf{Value (RoboMimic)} \\ \midrule
Hidden Size             & 256                    & 400                        \\
Number of Layers        & 4                      & 2                          \\
Number of Heads         & 8                      & 8                          \\
Pointwise Ratio         & 4                      & 4                          \\ \bottomrule
\end{tabular}
\label{supp:table:xf_xl}
\end{table}

\subsection{Action Decoding}
To decode joystick actions in DMLab tasks, we learn a 3-layer MLP whose output directly parameterizes a categorical distribution. This action head has a hidden dimension of 128 with ReLU activations. 
The ``Goal Maze'' and ``Irreversible Path'' tasks have an action dimension of 7, while ``Watermaze'' has 15 actions.
To decode continuous actions in RoboMimic, we learn a 2-layer MLP that parameterizes a Gaussian Mixture Model (GMM) with $5$ modes that generates a 7-dimensional action.
This network has a hidden dimension of 400 with ReLU activations.
During deployment, we employ the ``low-noise evaluation'' trick~\citep{hoffman2020acme}.

\section{Training Details and Hyperparameters}
\label{supp:sec:hyperparameters}

All experiments are conducted on cluster nodes with NVIDIA V100 GPUs.
We utilize DDP (distributed data parallel) to accelerate the training if necessary.
Training hyperparameters are listed in Table~\ref{supp:table:train_hyperparams}.

\begin{table}[!htbp]
\caption{Hyperparameters used during training.}
\vspace{0.1in}
\centering
\begin{tabular}{@{}lll@{}}
\toprule
\textbf{Hyperparameter}   & \textbf{Value (DMLab)} & \textbf{Value (RoboMimic)} \\ \midrule
Learning Rate             & 0.0005                 & 0.0001                     \\
Warmup Steps              & 1000                   & 0                          \\
LR Cosine Annealing Steps & 100000                 & N/A                        \\
Weight Decay              & 0.0                    & 0.0                        \\
\bottomrule                       
\end{tabular}
\label{supp:table:train_hyperparams}
\end{table}
\section{Experiment Details}
\label{supp:sec:experiment_details}

\subsection{DMLab Main Experiment}
Our DMLab main experiment is conducted on three levels with task IDs
\begin{itemize}
    \item{\texttt{explore\textunderscore{}goal\textunderscore{}locations\textunderscore{}large},}
    \item{\texttt{rooms\textunderscore{}watermaze},}
    \item{and \texttt{skymaze\textunderscore{}irreversible\textunderscore{}path\textunderscore{}hard}.}
\end{itemize}
We use no action repeats during training and evaluation.
For experiments with varying task difficulty, we select difficulty parameters ``room numbers'', ``spawn radius'', and ``built-in difficulty'' for these three levels, respectively.
We adopt environment wrappers and helper functions from \citet{petrenko2020sample} to flexibly and precisely maneuver task difficulties. 

Due to different task horizons, we tune the context length of Transformer-XL models and vary curricular trajectories accordingly. These differences are summarized in Table~\ref{supp:table:dmlab_main}.

\begin{table}[!htbp]
\caption{Experiment details on DMLab tasks. Columns ``Epoch'' denote the exact training epochs with best validation performance. We select these checkpoints for evaluation.
For task-difficulty-based curriculum, the column ``Training Trajectories'' with $n \times m$ entries means $n$ trajectories per difficulty level ($m$ levels in total). The column ``Sampled Episodes'' with $[i, j]$ entries means we first determine the number of episodes per difficulty level by uniformly sampling an integer from $[i, j]$ (inclusively).}
\centering
\vspace{0.1in}
\resizebox{1\textwidth}{!}{
\begin{tabular}{c|c|ccc|ccc}
\hline
\multirow{2}{*}{\textbf{\begin{tabular}[c]{@{}c@{}}Level\\ Name\end{tabular}}} & \multirow{2}{*}{\textbf{\begin{tabular}[c]{@{}c@{}}Context\\ Length\end{tabular}}} & \multicolumn{3}{c|}{\textbf{Task-Difficulty-Based Curriculum}}     & \multicolumn{3}{c}{\textbf{Learning-Progress-Based Curriculum}}    \\ \cline{3-8} 
                                                                               &                                                                                    & Epoch                   & Training Trajectories & Sampled Episodes & Epoch                   & Training Trajectories & Sampled Episodes \\ \hline
Goal Maze                                                                      & 500                                                                                & \multicolumn{1}{c|}{84} & 100 x 3               & {[}1, 5{]}       & \multicolumn{1}{c|}{88} & 300                   & 9                \\
Watermaze                                                                      & 400                                                                                & \multicolumn{1}{c|}{89} & 100 x 3               & {[}1, 5{]}       & \multicolumn{1}{c|}{80} & 300                   & 9                \\
Irreversible Path                                                              & 1600                                                                               & \multicolumn{1}{c|}{90} & 100 x 4               & {[}1, 3{]}       & \multicolumn{1}{c|}{97} & 400                   & 8                \\ \hline
\end{tabular}
}
\label{supp:table:dmlab_main}
\end{table}

RL oracles serve as source agents used to generate training data for our methods and the ``BC w/ Expert Data'' baseline.
They are trained with the PPO~\citep{schulman2017proximal} implementation from \citet{petrenko2020sample}.
The ``BC w/ Expert Data'' baselines have the same model architecture, training hyperparameters, and amount of training data as our method, but are trained solely on trajectories generated by the best performing RL oracles without cross-episodic attention.

\begin{table}[t!]
\caption{Evaluation results on DMLab, averaged over three tasks (Figure~\ref{fig:dmlab_core}).}
\label{supp:table:dmlab_main_avg_results}
\vspace{0.1in}
\centering
\resizebox{\textwidth}{!}{
\begin{tabular}{@{}cccccccccc@{}}
\toprule
\textbf{\begin{tabular}[c]{@{}c@{}}Ours (Task\\ Difficulty), Auto\end{tabular}} & \textbf{\begin{tabular}[c]{@{}c@{}}Ours (Task\\ Difficulty), Fixed\end{tabular}} & \textbf{\begin{tabular}[c]{@{}c@{}}Ours (Learning\\ Progress)\end{tabular}} & \textbf{\begin{tabular}[c]{@{}c@{}}DT (Mixed\\ Difficulty)\end{tabular}} & \textbf{\begin{tabular}[c]{@{}c@{}}DT (Single\\ Difficulty)\end{tabular}} & \textbf{\begin{tabular}[c]{@{}c@{}}AT (Mixed\\ Difficulty)\end{tabular}} & \textbf{\begin{tabular}[c]{@{}c@{}}AT (Single\\ Difficulty)\end{tabular}} & \textbf{\begin{tabular}[c]{@{}c@{}}BC w/ Expert\\ Data\end{tabular}} & \textbf{\begin{tabular}[c]{@{}c@{}}RL\\ (Oracle)\end{tabular}} & \textbf{\begin{tabular}[c]{@{}c@{}}Curriculum RL\\ (Oracle)\end{tabular}} \\ \midrule
51.4                                                                            & $\bestscore{54.4}$                                                                             & 32.4                                                                        & 35.3                                                                     & 11.7                                                                      & 42.7                                                                     & 33.4                                                                      & 14.2                                                                 & 40.6                                                           & 50.6                                                                      \\ \bottomrule
\end{tabular}
}
\end{table}

\subsection{DMLab Generalization}
\label{supp:sec:dmlab_generalization}
\begin{figure}[ht]
    \centering
    \includegraphics[scale=0.1]{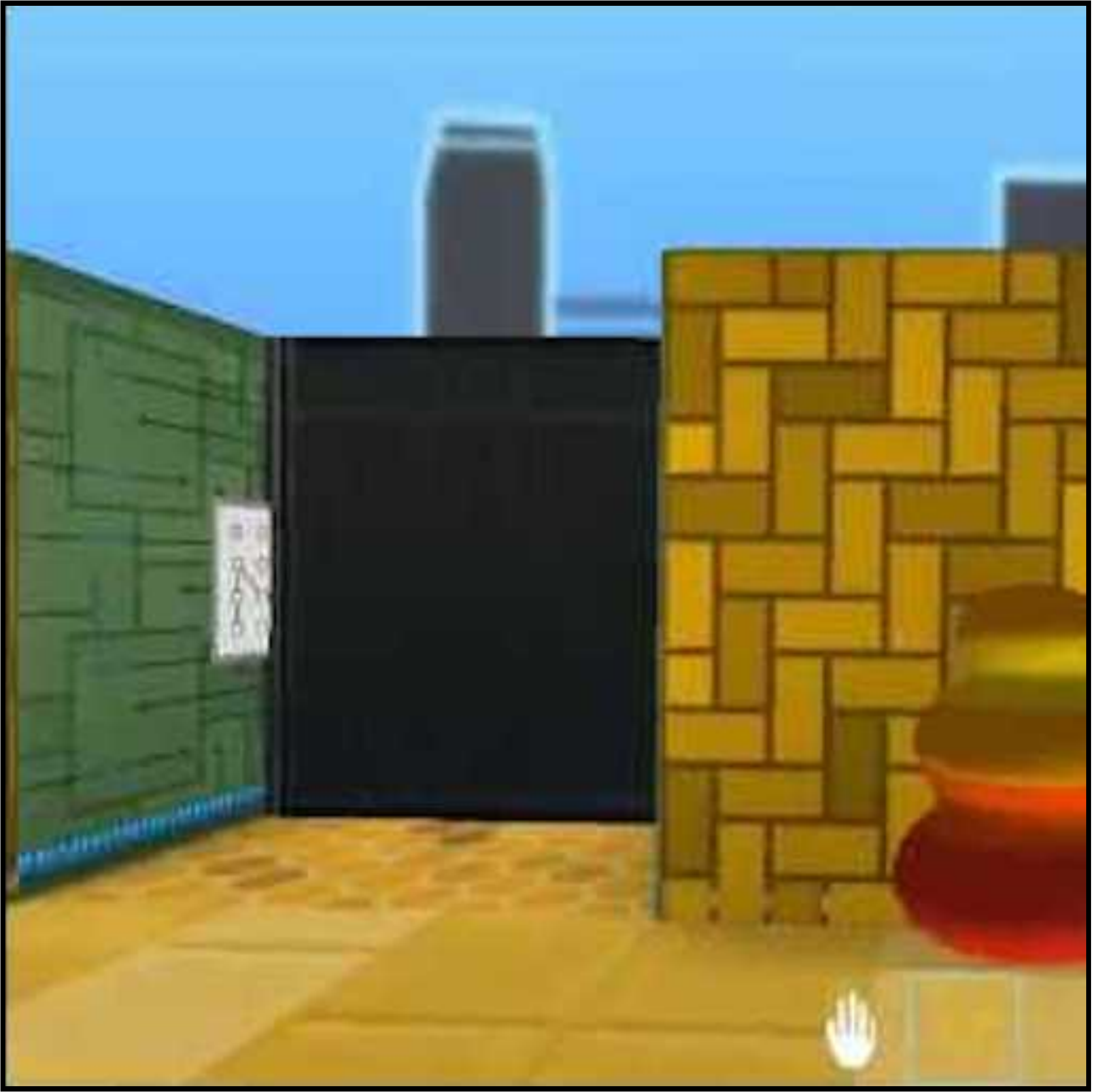}
    \caption{\textbf{A visualization of the task ``Goal Maze (Unseen Mechanism)''.} It includes doors that are randomly opened or closed.}
    \label{supp:fig:obstructed_goal_maze}
\end{figure}

This series of experiments probe the zero-shot generalization capabilities of embodied agents in unseen maze configurations, out-of-distribution difficulty levels, and varying environment dynamics.
For the task ``Goal Maze w/ Unseen Mechanism'', we use the level with task ID \texttt{explore\textunderscore{}obstructed\textunderscore{}goals\textunderscore{}large}, which adds randomly opened and closed doors into the maze while ensuring a valid path to the goal always exists.
An example of an agent's ego-centric observation is visualized in Figure~\ref{supp:fig:obstructed_goal_maze}.

The task ``Irreversible Path (OOD. Difficulty)'' corresponds to configurations with the built-in difficulty of 1 (agents are only trained on difficulty up to 0.9, as noted in Table~\ref{table:dmlab_exp_setting}).
For tasks with varying environment dynamics, we directly test agents with an action repeat of 2. This is different from the training setting with no action repeat.

\begin{table}[t!]
\caption{Generalization results on DMLab, averaged over five settings (Figure~\ref{fig:dmlab_generalization}).}
\label{supp:table:dmlab_gen_avg_results}
\vspace{0.1in}
\centering
\resizebox{\textwidth}{!}{
\begin{tabular}{@{}ccccccccc@{}}
\toprule
\textbf{\begin{tabular}[c]{@{}c@{}}Ours (Task\\ Difficulty)\end{tabular}} & \textbf{\begin{tabular}[c]{@{}c@{}}Ours (Learning\\ Progress)\end{tabular}} & \textbf{\begin{tabular}[c]{@{}c@{}}DT (Mixed\\ Difficulty)\end{tabular}} & \textbf{\begin{tabular}[c]{@{}c@{}}DT (Single\\ Difficulty)\end{tabular}} & \textbf{\begin{tabular}[c]{@{}c@{}}AT (Mixed\\ Difficulty)\end{tabular}} & \textbf{\begin{tabular}[c]{@{}c@{}}AT (Single\\ Difficulty)\end{tabular}} & \textbf{\begin{tabular}[c]{@{}c@{}}BC w/ Expert\\ Data\end{tabular}} & \textbf{\begin{tabular}[c]{@{}c@{}}RL\\ (Oracle)\end{tabular}} & \textbf{\begin{tabular}[c]{@{}c@{}}Curriculum RL\\ (Oracle)\end{tabular}} \\ \midrule
$\bestscore{39.6}$                                                                      & 27.8                                                                        & 31.8                                                                     & 13.6                                                                      & 39.4                                                                     & 29.2                                                                      & 18.1                                                                 & 30.0                                                           & 37.6                                                                      \\ \bottomrule
\end{tabular}
}
\end{table}

\subsection{RoboMimic Main Experiment}
\label{supp:sec:robomimic_main}

We leverage the Multi-Human (MH) dataset from \citet{mandlekar2021matters}. It consists of demonstrations collected by operators with varying proficiency.
We construct the expertise-based curriculum by following the order of ``worse operators, okay operators, then better operators''.
We use a context length of 200 for both tasks. There are 90 trajectories per expertise level. To determine the number of trajectories per expertise level when constructing curricular data, we uniformly sample an integer from $[1, 5]$ (inclusively).
The ``Lift'' and ``Can'' tasks are solved after training for 33 epochs and 179 epochs, respectively.
We control for the same number of training epochs in subsequent ablation studies.

\subsection{Ablation Study on Curriculum Granularity}
\label{supp:sec:curriculum_granularity}

We perform this ablation with the task-difficulty-based curriculum on DMLab levels due to the ease of adjusting granularity.
The definition of varying levels of curriculum coarseness is listed in Table~\ref{supp:table:curriculum_granularity}.

\begin{table}[!htbp]
\caption{Definitions of varying levels of curriculum coarseness.}
\vspace{0.1in}
\centering
\begin{tabular}{@{}cccccc@{}}
\toprule
\textbf{Level Name} & \textbf{Difficulty Parameter} & \textbf{Test Difficulty} & \textbf{Fine} & \textbf{Medium} & \textbf{Coarse} \\ \midrule
Goal Maze           & Room Numbers                  & 20                       & 5→10→15       & 5→10            & 5→15            \\
Watermaze           & Spawn Radius                  & 580                      & 150→300→450   & 150→300         & 150→450         \\
Irreversible Path   & Built-In Difficulty           & 0.9                      & .1→.3→.5→.7   & .1→.5→.7        & .1→.3→.5        \\ \bottomrule
\end{tabular}
\label{supp:table:curriculum_granularity}
\end{table}

\subsection{Comparison of Curricula in RoboMimic}
\label{supp:sec:curriculum_comparison}

In IL settings, we further explored the efficacy of various curricula. For the RoboMimic tasks examined, we employed a learning-progress-based curriculum, ensuring the total training trajectories matched those of the expertise-based curriculum (i.e., 270 trajectories per task). All other parameters remained consistent, with the training data derived from RoboMimic's machine-generated dataset.

Table~\ref{supp:table:curriculum_comparison} indicates that when heterogeneous-quality human demonstrations are accessible, the expertise-based curriculum is preferable due to its superior performance over the learning-progress-based approach. Conversely, without expert demonstrations and relying solely on machine-generated data, the learning-progress-based curriculum is still commendable. It offers noteworthy results and surpasses offline RL methods like CQL~\citep{kumar2020conservative}, even though CQL is trained on the full RoboMimic dataset, encompassing 1500 trajectories for the Lift task and 3900 for the Can task.

\begin{table}[t!]
\caption{Results show the performance of different curricula on two robotic manipulation tasks: Lift and Can. Standard deviations are included.}
\label{supp:table:curriculum_comparison}
\vspace{0.1in}
\centering
\resizebox{\textwidth}{!}{
\begin{tabular}{@{}lccc@{}}
\toprule
\textbf{Task} & \textbf{Expertise-Based Curriculum} & \textbf{Learning-Progress-Based Curriculum} & \textbf{CQL}~\citep{kumar2020conservative} \\ \midrule
Lift          &  $100.0 \pm 0.0$      &    $32.0 \pm 17.0$                          & $2.7 \pm 0.9$ \\
Can           &  $100.0 \pm 0.0$       &    $30.0 \pm 2.8$                           & $0.0 \pm 0.0$ \\
Average       &  $\bestscore{100.0}$               &    $31.0$                                  & $1.4$         \\ \bottomrule
\end{tabular}
}
\end{table}

\section{Feasibility of Obtaining Curricular Data}
\label{supp:sec:feasibility}

The challenge of accurately orchestrating a curriculum is non-trivial and hinges on various factors. In the present work, three curriculum designs are introduced and validated, each with its practical considerations and underlying assumptions, discussed herein.

\para{Learning-Progress-Based Curriculum.}
RL agents typically exhibit monotonic improvement over training epochs, thereby naturally producing incrementally better data. The curriculum here is devised through a series of checkpoints throughout the training duration, necessitating no supplementary assumptions for its formulation.

\para{Task-Difficulty-Based Curriculum.}
In contexts where environmental difficulty is parameterizable, curricula can be structured through a schedule, determined by the relevant difficulty parameter, as demonstrated within this work. In scenarios lacking parameterized difficulty, alternatives such as methods proposed by \citet{kanitscheider2021multitask} may be employed. The application of our method to tasks where difficulty is not explicitly characterized presents an intriguing avenue for future research.

\para{Expertise-Based Curriculum.}
A notable limitation resides in the requisite to estimate demonstrators’ proficiency. While some IL benchmarks, e.g., RoboMimic \citep{mandlekar2021matters}, come pre-equipped with proficiency labels, a broader application of our method necessitates an approximation of proficiency. One plausible approach entails ranking trajectories via completion time. Furthermore, a demonstrator's proficiency is likely to organically improve—from initial unfamiliarity with teleoperation systems or tasks, to a stage of executing data collection with muscle memory \citep{mandlekar2018roboturk}. This progression potentially provides a rich learning signal conducive for CEC application.
\section{Broader Impact}
\label{supp:sec:broader_impact}

Our Cross-Episodic Curriculum can significantly enhance Transformer agent learning but carries potential societal impacts. 
The efficiency of our method depends on the curriculum's design. If the curriculum unintentionally reflects biases, it could lead to the amplification of these biases in learned policies, potentially perpetuating unfair or discriminatory outcomes in AI-driven decisions.
Furthermore, the computational intensity of our approach at evaluation could contribute to increased energy usage, which has implications for the environmental footprint of AI applications.

\end{document}